\documentclass{article} 
\usepackage[final]{colm2026_conference}

\usepackage{microtype}
\usepackage{hyperref}
\usepackage{url}
\usepackage{lineno}

\usepackage[T1]{fontenc}
\usepackage[utf8]{inputenc}

\usepackage{amsmath,amssymb,amsfonts,amsthm}
\usepackage{subcaption}
\usepackage{booktabs}
\usepackage{multirow}
\usepackage{graphicx}
\usepackage{pifont}
\usepackage{etoc}
\usepackage{cleveref}
\crefformat{figure}{Fig.~#2#1#3}
\crefformat{theorem}{Theorem~#2#1#3}
\crefformat{lemma}{Lemma~#2#1#3}
\crefformat{algorithm}{Alg.~#2#1#3}
\crefformat{section}{Sec.~#2#1#3}
\crefformat{equation}{Eq.~#2#1#3}

\definecolor{darkblue}{rgb}{0, 0, 0.5}
\hypersetup{colorlinks=true, citecolor=darkblue, linkcolor=darkblue, urlcolor=darkblue}

\title{ReflCtrl: Controlling LLM Reflection Efficiently via \\ Representation Engineering}

\author{Ge Yan \\
CSE, UCSD \\
\texttt{geyan@ucsd.edu} \And 
Chung-En Sun \\
CSE, UCSD \\
\texttt{cesun@ucsd.edu} \And
Linbo Liu \\
UCSD \\
\texttt{linbo@ucsd.edu}\And 
Tsui-Wei (Lily) Weng \\
HDSI, UCSD \\
\texttt{lweng@ucsd.edu}
}

\newcommand{\blfootnote}[1]{%
  \begingroup
  \renewcommand\thefootnote{}\footnote{#1}%
  \addtocounter{footnote}{-1}%
  \endgroup
}

\begin{document}

\ifcolmsubmission
\linenumbers
\fi

\maketitle
\blfootnote{Code: \url{https://github.com/Trustworthy-ML-Lab/ReflCtrl}}
\blfootnote{Project website: \url{https://lilywenglab.github.io/ReflCtrl/}}
\begin{abstract}
Large reasoning models achieve strong performance on diverse tasks by producing extended chains of thought. \textbf{Self-reflection}, the ability to review and revise prior reasoning steps, is widely regarded as a key contributor to this performance. However, self-reflection also incurs substantial inference cost, and its governing mechanism remains underexplored. In this work, we study self-reflection through the lens of representation engineering. First, we identify a \textit{reflection direction} in the model's latent space that separates reflection steps from non-reflection steps, and show that activation along this direction is strongly predictive of answer correctness, suggesting that self-reflection is regulated by the model's internal uncertainty. Next, building on this insight, we propose \textbf{ReflCtrl}, a framework that controls self-reflection via a \textit{stepwise steering method}: interventions are applied only at the start of each new reasoning step, enabling fine-grained control over reflection frequency without degrading generation quality. Experiments across math and general reasoning benchmarks show that reflection is often redundant, especially in stronger models: ReflCtrl reduces total reasoning tokens by up to 43.2\% while preserving accuracy, and substantially outperforms the conventional approach that steers at every token, at matched token budgets.
\end{abstract}

\etocdepthtag.toc{main}
\section{Introduction}
\label{sec:intro}
Reasoning language models, such as OpenAI o1~\citep{openai2024o1} and DeepSeek-R1~\citep{deepseekai2025deepseekr1incentivizingreasoningcapability}, achieve strong performance on diverse tasks including mathematics, coding, and general reasoning by producing extended chains of thought before a final answer. A distinctive behavior of these models, emerging from training rather than explicit prompting, is \textbf{self-reflection}: the ability to spontaneously review and revise prior reasoning steps during inference. While self-reflection is widely regarded as a key contributor to reasoning quality, it also represents a substantial inference overhead: our analysis shows reflection steps account for 23--36\% of total reasoning tokens (see \cref{fig:reflTokenPie}), varying with benchmark difficulty.

\begin{figure*}[tp]
    \centering
    \includegraphics[width=0.95\linewidth]{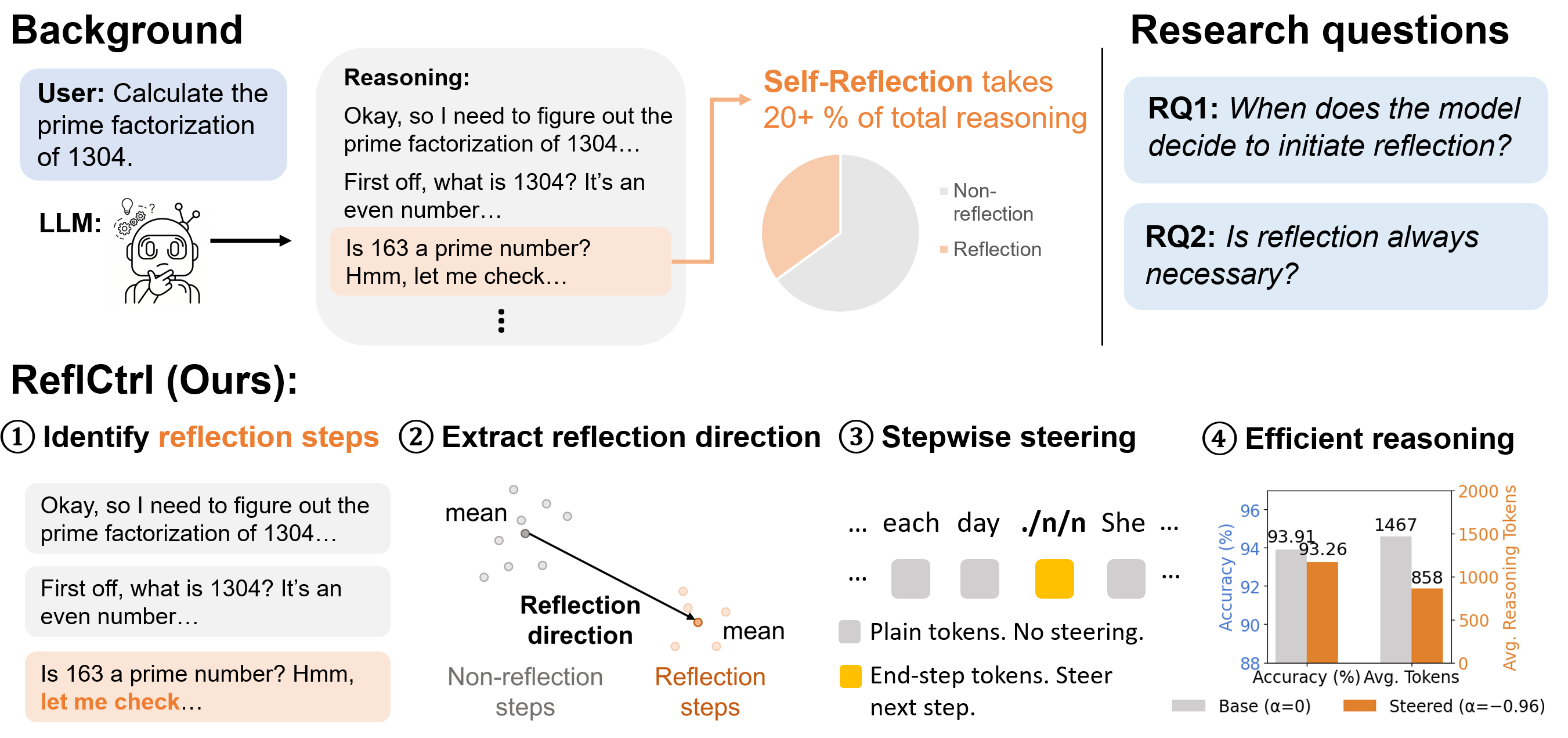}
    \caption{\textbf{Overview of the proposed ReflCtrl framework.} (1) Reflection steps are identified from the model's chain-of-thought reasoning. (2) A reflection direction is extracted by computing the mean difference between reflection and non-reflection step representations in the latent space. (3) During inference, steering is applied only at end-of-step tokens (e.g., \texttt{\textbackslash n\textbackslash n}), leaving plain tokens unmodified. (4) This stepwise steering reduces reasoning tokens with minimal accuracy loss.}
    \label{fig:overview}
\end{figure*}

Despite this prevalence and cost, the mechanism governing self-reflection is less explored. Two fundamental questions remain open. 
\begin{enumerate}
    \item \textbf{(RQ1)} \emph{When does the model decide to initiate reflection?} Without identifying the internal trigger, it is unclear whether reflection is a principled response to uncertainty or a spurious training artifact.
    \item \textbf{(RQ2)} \emph{Is reflection always necessary?} If many reflection steps are redundant, suppressing them could substantially reduce inference cost, but only with a reliable mechanism to identify and control reflection.
\end{enumerate}
Answering these questions requires a way to probe self-reflection in the model's internal representations, which prior work has not provided.
In this work, we address both questions through the lens of \textbf{representation engineering}~\citep{zou2023representation}. We propose \textbf{ReflCtrl}, a framework that identifies a \textit{reflection direction} in the model's latent space: a direction that separates reflection steps from non-reflection steps in hidden-state activations (see \cref{fig:overview}). Specifically, we segment the model's reasoning trace into steps delimited by natural paragraph breaks, identify reflection-related steps via keywords, and compute the reflection direction as the mean difference in hidden representations between the two sets. Using this direction, we introduce a \textit{stepwise steering method} that injects the direction only at the start of each new reasoning step, enabling fine-grained and efficient control over reflection frequency without degrading generation quality.

Our experiments yield two key findings. First, model activations along the reflection direction are highly predictive of answer correctness, indicating that self-reflection is internally regulated by the model's uncertainty and providing a mechanistic answer to \textbf{RQ1}. Second, reflection is often redundant: applying stepwise steering to suppress unnecessary reflections reduces total reasoning tokens by up to 43.2\% while preserving reasoning accuracy, directly addressing \textbf{RQ2}. Together, these results offer both a mechanistic account of self-reflection and a practical method for inference-time efficiency.

Our contributions are:
\begin{enumerate}
    \item We propose \textbf{ReflCtrl}, a framework that identifies a reflection direction in LLM latent space and uses it to control self-reflection via stepwise representation steering.
    \item We show that activations along the reflection direction correlate with model uncertainty, providing a mechanistic explanation for when and why models initiate self-reflection (\textbf{RQ1}).
    \item We demonstrate that many reflections are redundant and that suppressing them via stepwise steering reduces inference cost by up to 43.2\% without sacrificing reasoning accuracy (\textbf{RQ2}).
\end{enumerate}
\section{Preliminaries}
\label{sec:prelim}

\subsection{Background}
Reasoning LLMs are built upon the Transformer decoder architecture~\citep{vaswani2017attention}.
Our method extracts hidden representations from two sub-components of each layer: the self-attention block and the feed-forward MLP block.
Each decoder layer $l$ processes the hidden representation $z_l \in \mathbb{R}^d$ as:
\begin{equation}
\begin{aligned}
    &\tilde{z}_l = z_l + z^{\text{attn}}_l, \;z^{\text{attn}}_l = \text{Attn}(\text{LN}(z_l)),\\
    & z_{l+1} = \tilde{z}_l + z^{\text{mlp}}_l, \;z^{\text{mlp}}_l = \text{MLP}(\text{LN}(\tilde{z}_l)).
\end{aligned}
\end{equation}
Here, $\text{LN}(\cdot)$ denotes layer normalization, $\text{Attn}(\cdot)$ is the self-attention block, and $\text{MLP}(\cdot)$ is the feed-forward network.
We denote $\tilde{z}_l$ as the intermediate state after the attention block, and use $z_l^{\text{attn}}$ and $z_l^{\text{mlp}}$ as the per-step representations from which we extract the reflection direction.

\subsection{Identifying reflection steps}
\label{subsec:identify}
Reasoning LLMs generate extended chain-of-thought traces that interleave forward reasoning with \emph{self-reflection} \citep{deepseekai2025deepseekr1incentivizingreasoningcapability}: ``aha moments'' at which the model pauses to re-examine, critique, or revise its prior reasoning before proceeding.
These self-reflective steps are typically surfaced through characteristic phrases such as ``Wait'' or ``Let me check'', signaling that the model has shifted from producing new reasoning content to evaluating what it has already produced.

To identify self-reflection in model outputs, we employ a two-stage process. First, we segment each reasoning trace into individual steps by leveraging the consistent delimiter pattern used by reasoning LLMs: the token sequence ``\textbackslash n\textbackslash n'' (empty lines). This natural segmentation produces coherent reasoning chunks that serve as our basic unit of analysis.
Second, we classify each chunk as either forward reasoning or self-reflection using a keyword-based approach. We search for characteristic phrases that signal the initiation of reflective episodes, such as ``Wait'' and ``Let me check''. This heuristic captures the model's explicit transitions from generating new reasoning content to evaluating previously generated steps. We validate the reliability of this labeling scheme using an LLM-based judge (\cref{app:reflection-audit}).

\subsection{Empirical motivation for causal control}
\begin{figure}[!htbp]
    \centering
    \begin{subfigure}{0.48\linewidth}
        \centering
        \includegraphics[width=\linewidth]{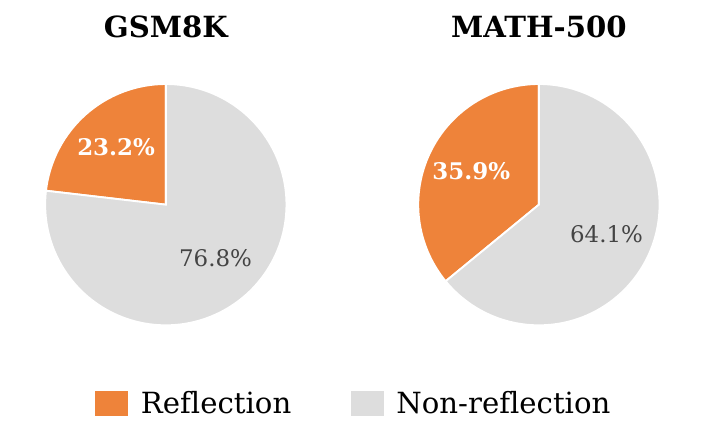}
        \caption{Reflection token proportion}
        \label{fig:reflTokenPie}
    \end{subfigure}
    \hfill
    \begin{subfigure}{0.48\linewidth}
        \centering
        \includegraphics[width=\linewidth]{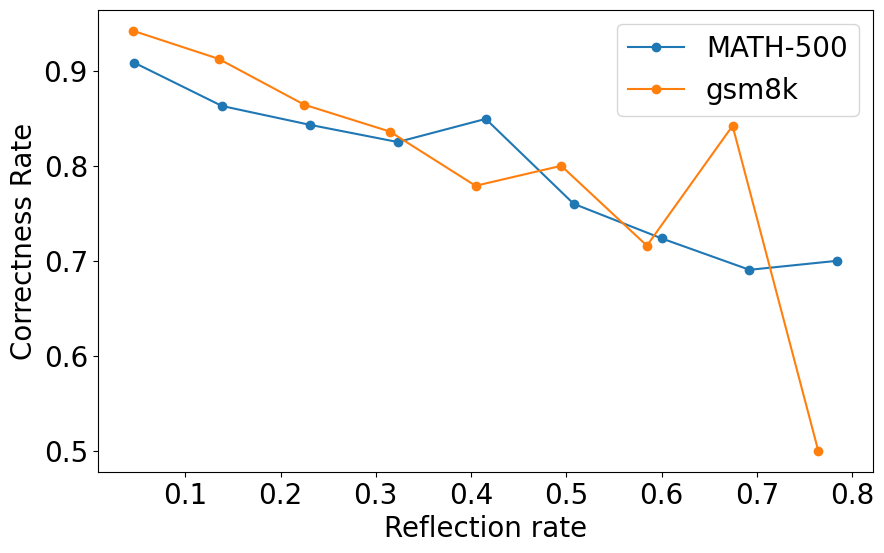}
        \caption{Correctness vs.\ reflection rate}
        \label{fig:accReflRate}
    \end{subfigure}
    \caption{\textbf{Reflection overhead and its correlation with accuracy.} \textit{Left}: Reflection steps account for 23.2\% and 35.9\% of all reasoning tokens on GSM8k and MATH-500, respectively, for DeepSeek-R1-Distill-Llama-8B. \textit{Right}: Higher reflection frequency is associated with lower accuracy on both benchmarks ($N=10$ samples per question), but this trend is confounded by question difficulty (see \cref{app:verify}).}
    \label{fig:reflMotivation}
\end{figure}

Self-reflection, while potentially valuable, introduces non-trivial inference overhead.
\Cref{fig:reflTokenPie} quantifies this cost: on the DeepSeek-R1-Distill-Llama-8B baseline, reflection steps consume \textbf{23.2\%} of all reasoning tokens on GSM8k and \textbf{35.9\%} on MATH-500.

To investigate \textbf{RQ2}, we examine how accuracy relates to reflection frequency.
Using DeepSeek-R1-Distill-Llama-8B on GSM8k and MATH-500 with $N=10$ samples per question, we find that correctness rate decreases as reflection rate (\# of reflection steps / \# of total reasoning steps) increases (\cref{fig:accReflRate}). This may seem surprising at the first glance, as reflection appears to hurt reasoning performance. However, note that the correlation does \emph{not} imply reflection is useless: harder questions usually trigger more reflections. Since those questions have lower accuracy, the reflection frequency is confounded with task difficulty (see \cref{app:verify} for deeper analysis).

To disentangle the causal effect of reflection from the confounder of question difficulty, we need to manipulate reflection behavior for the same question.
This inspires us to propose \textbf{ReflCtrl}: a representation engineering approach that identifies this \emph{reflection direction} from labeled hidden states and steers it at inference time to causally control reflection frequency.

\section{ReflCtrl: Probing and steering self-reflection}
\label{sec:method}
In this section, we propose ReflCtrl, a representation engineering pipeline that enables flexible, inference-time control of the model's self-reflection without degrading normal generation.
The core idea is to identify a \textit{reflection direction}---a direction in the model's latent space that linearly separates reflection steps from non-reflection steps---and use it both to steer reflection frequency during inference (\cref{subsec:steer}) and to probe the model's internal state as an approach to \textbf{RQ1} (\cref{sec:uqRefl}).
The method proceeds in two stages: computing the reflection direction from labeled hidden states (\cref{subsec:extract}), and applying stepwise representation steering to control self-reflection (\cref{subsec:steer}).

\subsection{Extracting the reflection direction}
\label{subsec:extract}
Given the labeled reflection and non-reflection steps from \cref{subsec:identify}, we compute the reflection direction using the mean-difference method from representation engineering~\citep{zou2023representation}.
For each step $s$ at layer $l$, we extract the hidden representations from the attention and MLP outputs at the \textit{first token} of the step, denoted $z_l^{\{\text{attn,mlp}\}}(s)$.
We choose the first token because it captures the model's internal state at the moment a new reasoning step is initiated, which is the position where the decision to reflect or continue forward reasoning is made.
The reflection direction is then the mean difference between reflection and non-reflection representations:
\begin{equation}
    d_l^{\{\text{attn,mlp}\}} = \frac{1}{|\text{R}|}\sum_{s \in \text{R}}z_l^{\{\text{attn,mlp}\}}(s) 
    - \frac{1}{|\text{NR}|}\sum_{s \in \text{NR}}z_l^{\{\text{attn,mlp}\}}(s),
\end{equation}
where $\text{R}$ and $\text{NR}$ are the sets of reflection and non-reflection steps, respectively.
The mean-difference estimator is simple, interpretable, and well-motivated by the linear representation hypothesis~\citep{zou2023representation}: it produces a single direction that maximally separates the two classes in expectation, without requiring a learned classifier or labeled supervision.

\subsection{Stepwise steering}
\label{subsec:steer}
The extracted reflection direction enables inference-time control: adding it to internal representations steers the model toward or away from reflection.
The standard additive intervention at layer $l$ modifies the hidden state as:
\begin{equation}
    z_{l, \text{intv}}^{\{\text{attn,mlp}\}} = z_l^{\{\text{attn,mlp}\}} + \lambda \, d_l^{\{\text{attn,mlp}\}},
    \label{eq:intervention}
\end{equation}
where $\lambda \in \mathbb{R}$ is the intervention strength: $\lambda < 0$ suppresses reflection and $\lambda > 0$ promotes it.

In prior representation steering work~\citep{zou2023representation}, this intervention is applied at every token position during autoregressive generation.
Let $(x_1, x_2, \ldots, x_T)$ denote the generated token sequence.
The \textit{all-token} strategy applies \cref{eq:intervention} uniformly:
\begin{equation}
    z_{l, \text{intv}}^{\{\text{attn,mlp}\}}(x_t) = z_l^{\{\text{attn,mlp}\}}(x_t) + \lambda \, d_l^{\{\text{attn,mlp}\}}, \;\; \forall\, t \in \{1, \ldots, T\}.
    \label{eq:alltoken}
\end{equation}
Because generation is autoregressive, each perturbed hidden state feeds into subsequent attention computations, causing the distributional shift to compound across tokens.
At high $|\lambda|$, this cumulative effect pushes the model's representations far from the training distribution, degrading generation quality and limiting the usable intervention range.

To address this, we propose \textbf{stepwise steering}: instead of intervening at every token, we restrict the perturbation to \textit{step-boundary positions}, the first token generated after each reasoning step delimiter.
Formally, let $\mathcal{B} = \{t : x_{t-1} \text{ completes the delimiter ``\textbackslash n\textbackslash n''}\}$ denote the set of step-boundary positions.
The stepwise intervention replaces \cref{eq:alltoken} with:
\begin{equation}
    z_{l, \text{intv}}^{\{\text{attn,mlp}\}}(x_t) = z_l^{\{\text{attn,mlp}\}}(x_t) + \lambda \, \mathbf{1}[t \in \mathcal{B}] \, d_l^{\{\text{attn,mlp}\}},
    \label{eq:stepwise}
\end{equation}
where $\mathbf{1}[\cdot]$ is the indicator function.
Since the number of reasoning steps $|\mathcal{B}|$ is typically much smaller than the total token count $T$ (e.g., a 2000-token trace may contain only 10--20 steps), stepwise steering drastically reduces the number of perturbed positions.

This design is motivated by two observations.
First, the reflection direction is extracted from first-token representations (\cref{subsec:extract}), so intervening at step boundaries targets the positions where the direction is most aligned with the model's step-initiation mechanism.
Second, leaving intermediate tokens within each step unperturbed preserves the model's natural generation dynamics, avoiding the compounding distributional drift that degrades all-token steering at high $|\lambda|$.
As shown in \cref{fig:stepbeginFull,fig:stepbeginToken}, stepwise steering maintains the same direction of effect as all-token steering while supporting a substantially wider range of $\lambda$, enabling finer-grained control over reflection frequency without sacrificing generation quality.

\section{Experiments}
In this section, we present experiments evaluating our ReflCtrl framework on three fronts: what triggers self-reflection (\textbf{RQ1}), how reflection influences model reasoning performance (\textbf{RQ2}), and how stepwise steering reduces inference cost in practice.
\subsection{Settings}
\paragraph{Models.} We primarily study the DeepSeek-R1-Distill series \citep{deepseekai2025deepseekr1incentivizingreasoningcapability}, including the Qwen2.5-14B and Llama-8B distilled variants, as these models are publicly available. We also evaluate QwQ-32B \citep{qwq-32b-preview} as a representative non-distilled reasoning model.
\paragraph{Datasets.} For math tasks, we use GSM8k \citep{cobbe2021gsm8k} and MATH-500 \citep{lightman2023let}  as benchmarks. For general reasoning, we evaluate on three MMLU \citep{hendrycks2020measuring} subsets: Professional Accounting, High School Computer Science, and Formal Logic.
\paragraph{Generation settings.} We follow the standard generation configuration for each model. For math tasks, we prompt the model with ``Please reason step by step, and put your final answer within \textbackslash boxed\{\}''. For MMLU, we use the same prompt and additionally require the final answer to be a single letter. Maximum completion tokens are set to 8192 for all tasks except MATH-500, where we use 16384 due to its higher complexity.

\paragraph{Reflection direction extraction.} We extract the reflection direction using GSM8k responses following the procedure in \cref{sec:method}. The final reasoning step is excluded, as it typically consists of a concluding sentence unrelated to the reasoning process.
\paragraph{Steering.} Unless otherwise specified, all results use the stepwise steering method proposed in \cref{sec:method}. Interventions are applied to all layers except the first and last six; we study the effect of this choice in \cref{app:ablationLayer}.
\subsection{Probing model uncertainty (RQ1)}
\label{sec:uqRefl}
We first use the reflection direction to address \textbf{RQ1}: \textit{when does the model decide to initiate reflection?}
We hypothesize that the model's internal uncertainty is the primary trigger: a reasoning LLM initiates reflection when it is uncertain about the current reasoning trajectory.
We conduct two experiments to verify this:
If this is correct, the reflection direction should encode information about answer correctness, since uncertainty and eventual correctness are tightly coupled.

\textbf{Experiment 1: }Correctness presiction: If the hypothesis is correct, the reflection direction should encode information about answer correctness, since uncertainty and eventual correctness are tightly coupled. We follow \citet{mielke-etal-2022-reducing} and train a logistic regression classifier to predict whether the model's answer is correct, using features derived from the reflection direction.
For each instance, we compute the projection of the hidden representation on the reflection direction at each layer and sub-component, and concatenate these into a feature vector $p_{\text{intv}}$:
\begin{equation}
    \begin{aligned}
       & p_{\text{intv}} = \text{concat}(\{{p_l^{\text{attn}}}\}_{l=1}^{N_{\text{layer}}}, \{{p_l^{\text{mlp}}}\}_{l=1}^{N_{\text{layer}}}),
    \end{aligned}
\end{equation}
where $p_l^{\{\text{attn}, \text{mlp}\}} = \frac{<d_l^{\{\text{attn}, \text{mlp}\}}, z_l^{\{\text{attn}, \text{mlp}\}}>}{\|z_l^{\{\text{attn}, \text{mlp}\}}\|_2}$ is the projection of representation.
We extract $p_{\text{intv}}$ from the end-of-thinking token (\textless/think\textgreater) and train the classifier on the GSM8k training set.
As a baseline, we train the same classifier using the last-layer embedding of the final token.

Results on the GSM8k test set (\cref{tab:probe_results}) show that reflection-direction features achieve higher AUROC and F1 than the final-layer baseline, despite having lower dimensionality.
This suggests that \textbf{model uncertainty is encoded in the reflection direction}, and provides evidence that uncertainty is a key mechanistic trigger for self-reflection in reasoning LLMs.
\begin{table}[!h]
\centering
\begin{tabular}{l|cc|cc}
\hline
\multirow{2}{*}{Model} & \multicolumn{2}{c|}{final layer embedding} & \multicolumn{2}{c}{reflection direction} \\
\cline{2-5}
& AUROC & F1 & AUROC & F1 \\
\hline
DS-Llama-8B & 0.736 & 0.946 & \textbf{0.772} & \textbf{0.948} \\
QwQ-32B & 0.555 & 0.636 & \textbf{0.564} & \textbf{0.839} \\
DS-Qwen-14B & 0.716 & 0.929 & \textbf{0.850} & \textbf{0.976} \\
\hline
\end{tabular}
\caption{\textbf{Probing results for uncertainty detection.} We train a logistic regression classifier to predict answer correctness using (i) the last token embedding at the final layer or (ii) feature vector $p_{\text{intv}}$ derived from reflection direction. Reflection-based features achieve higher AUROC and F1 scores despite lower dimensionality, suggesting that uncertainty is encoded in the reflection direction.}
\label{tab:probe_results}
\end{table}

\textbf{Experiment 2: Token-level entropy}
To strengthen this evidence, we compute per-token Shannon entropy (over the full vocabulary) at every generated position for all 1{,}319 GSM8k questions (DeepSeek-R1-Distill-Llama-8B) and compare the entropy of step $i$ conditioned on whether step $i{+}1$ is a reflection step.
Across 50{,}456 consecutive step pairs (16{,}582 pre-reflection, 33{,}874 pre-non-reflection), steps immediately preceding reflection exhibit significantly higher entropy (\cref{tab:entropy_analysis}) with both mean and max pooling, providing a direct evidence that uncertainty is an important factor in model's reflection.

\begin{table}[!h]
\centering
\begin{tabular}{l|cc|c}
\hline
Pooling & Pre-Reflect & Pre-Non-Reflect & Cohen's $d$ \\
\hline
Mean & 0.734 $\pm$ 0.332 & 0.599 $\pm$ 0.366 & 0.38 \\
Max & 2.401 $\pm$ 0.647 & 2.028 $\pm$ 0.789 & \textbf{0.50} \\
\hline
\end{tabular}
\caption{\textbf{Token entropy preceding reflection vs.\ non-reflection steps.} We compute per-token Shannon entropy at every output position and compare the entropy of step $i$ conditioned on whether step $i{+}1$ is a reflection step. $n{=}16{,}582$ pre-reflection and $n{=}33{,}874$ pre-non-reflection step pairs from DS-Llama-8B on GSM8k.}
\label{tab:entropy_analysis}
\end{table}

\subsection{Reflection redundancy (RQ2)}
Next, we continue to study \textbf{RQ2}: Reflection redundancy. As discussed in \cref{sec:prelim}, to understand the impact of reflections, we need to intervene on the model to change the number of reflections, and measure the resulting change in accuracy. For this purpose, we apply different intervention strengths and show results in \cref{tab:main_results}. Two key observations emerge:
\begin{enumerate}
    \item \textbf{Reflections are largely redundant in stronger models.} For QwQ-32B, suppressing reflections at strength $\lambda=-0.96$ reduces token usage by 32.4\% on GSM8k and 20.9\% on MATH-500, with accuracy drops of only 0.14pp and 0.34pp, respectively. Additionally, we note that the redundancy is more significant on larger models. To further isolate the effect of model scale from architecture differences, we evaluate three models from the same family (DeepSeek-R1-Distill-Qwen-1.5B/7B/14B) in \cref{app:same_family}, which further shows larger models are more resilient to the reduction of reflection, implying more redundancy.  
    \item \textbf{Smaller distilled models benefit only marginally from additional reflections.} DeepSeek-R1-Distill-Llama-8B gains only 0.16pp and 0.92pp in accuracy at strength $\lambda=+0.48$, at the cost of approximately 2000 additional tokens per question, which is an unfavorable trade-off in most scenarios.
\end{enumerate}
\begin{table*}[!t]
\centering
\resizebox{0.9\textwidth}{!}{%
\begin{tabular}{llccccc}
\toprule
\multirow{2}{*}{Dataset} & \multirow{2}{*}{Model} & \multirow{2}{*}{Metric} & \multicolumn{4}{c}{Intervention Strength $\lambda$} \\
\cmidrule(lr){4-7}
& & & $-0.96$ & $-0.48$ & $0$ & $+0.48$ \\
\midrule
\multirow{6}{*}{GSM8k}
 & DS-Llama-8B  & Acc (\%) $\uparrow$ & 88.34\;{\scriptsize(-1.75pp)} & 89.46\;{\scriptsize(-0.63pp)} & 90.09 & 90.25\;{\scriptsize(+0.16pp)} \\
 &              & Tokens $\downarrow$ & 821.0\;{\scriptsize(-48.5\%)} & \textbf{1032.6\;{\scriptsize(-35.3\%)}} & 1595.7 & 3577.1\;{\scriptsize(+124.2\%)} \\
\cmidrule(lr){2-7}
 & QwQ-32B  & Acc (\%) $\uparrow$ & 96.36\;{\scriptsize(-0.14pp)} & 96.50\;{\scriptsize(+0.00pp)} & 96.50 & 96.44\;{\scriptsize(-0.06pp)} \\
 &              & Tokens $\downarrow$ & \textbf{1006.7\;{\scriptsize(-32.4\%)}} & 1162.5\;{\scriptsize(-21.9\%)} & 1488.6 & 2256.9\;{\scriptsize(+51.6\%)} \\
\cmidrule(lr){2-7}
 & DS-Qwen-14B  & Acc (\%) $\uparrow$ & 95.07\;{\scriptsize(-0.08pp)} & 95.15\;{\scriptsize(+0.00pp)} & 95.15 & 94.84\;{\scriptsize(-0.31pp)} \\
 &              & Tokens $\downarrow$ & \textbf{747.8\;{\scriptsize(-43.2\%)}} & 880.2\;{\scriptsize(-33.1\%)} & 1315.9 & 3746.4\;{\scriptsize(+184.7\%)} \\
\midrule
\multirow{6}{*}{MATH-500}
 & DS-Llama-8B  & Acc (\%) $\uparrow$ & 82.14\;{\scriptsize(-3.84pp)} & 84.46\;{\scriptsize(-1.52pp)} & 85.98 & 86.90\;{\scriptsize(+0.92pp)} \\
 &              & Tokens $\downarrow$ & 2738.1\;{\scriptsize(-31.6\%)} & 3123.8\;{\scriptsize(-21.9\%)} & \textbf{4000.7} & 6017.8\;{\scriptsize(+50.4\%)} \\
\cmidrule(lr){2-7}
 & QwQ-32B  & Acc (\%) $\uparrow$ & 92.72\;{\scriptsize(-0.34pp)} & 92.58\;{\scriptsize(-0.48pp)} & 93.06 & 93.08\;{\scriptsize(+0.02pp)} \\
 &              & Tokens $\downarrow$ & \textbf{2992.9\;{\scriptsize(-20.9\%)}} & 3253.4\;{\scriptsize(-14.1\%)} & 3786.0 & 5028.9\;{\scriptsize(+32.8\%)} \\
\cmidrule(lr){2-7}
 & DS-Qwen-14B  & Acc (\%) $\uparrow$ & 89.22\;{\scriptsize(-2.22pp)} & 90.18\;{\scriptsize(-1.26pp)} & 91.44 & 91.86\;{\scriptsize(+0.42pp)} \\
 &              & Tokens $\downarrow$ & 2247.1\;{\scriptsize(-32.2\%)} & 2534.7\;{\scriptsize(-23.5\%)} & \textbf{3315.3} & 5789.0\;{\scriptsize(+74.6\%)} \\
\bottomrule
\end{tabular}
}
\caption{\textbf{Accuracy and average reasoning token usage under different intervention strengths.} Results on GSM8k and MATH-500 for three models, averaged over 10 runs. Negative intervention strengths consistently reduce token usage with minimal accuracy loss. Relative changes compared to the baseline ($\lambda=0$) are shown in parentheses. In the \emph{Tokens} ($\downarrow$) rows, the \textbf{bolded} token value (including its relative change) marks the lowest token usage (within each dataset/model pair) among intervention strengths whose \emph{accuracy drop} relative to the baseline is $<1$ percentage point (pp). Standard deviations are reported in \cref{app:std}.}
\label{tab:main_results}
\end{table*}

The full accuracy--token trade-off across all models and datasets is summarized in \cref{tab:main_results}. Results consistently show that reflection frequency can be substantially reduced with minimal accuracy loss, confirming reflection redundancy across model scales and benchmarks, especially for those larger and stronger models. 

\paragraph{Token savings across difficulty levels.}
A potential concern is whether the token reduction comes primarily from the model ``giving up'' on hard questions. To further understand the mechanism, we stratify MATH-500 questions into Easy ($>$80\% baseline accuracy, $n{=}399$), Medium (50--80\%, $n{=}41$), and Hard ($<$50\%, $n{=}60$) based on 10-sample per-question accuracy of DS-Llama-8B (\cref{tab:difficulty_stratification}). We have two observations:
(1) Token reduction is relatively uniform across all difficulty levels (20--33\%), not concentrated in hard questions. Instead, the reduction is \emph{larger} for easy questions (33.2\%) than hard ones (20.0\%).
(2) Regarding the accuraccy, hard-question accuracy is unchanged ($-$0.2pp), likely because the reflection does not help much for those questions. 

\begin{table}[!h]
\centering
\begin{tabular}{l|r|cc|c}
\hline
Difficulty & \#Q & $\lambda{=}0$ Acc & $\lambda{=}{-}0.96$ Acc & Token Red. \\
\hline
Easy & 399 & 98.0\% & 94.8\%\;{\scriptsize($-$3.2pp)} & 33.2\% \\
Medium & 41 & 60.0\% & 53.2\%\;{\scriptsize($-$6.8pp)} & 26.3\% \\
Hard & 60 & 12.5\% & 12.3\%\;{\scriptsize($-$0.2pp)} & 20.0\% \\
\hline
\end{tabular}
\caption{\textbf{Per-difficulty stratification on MATH-500} (DS-Llama-8B, $\lambda{=}{-}0.96$). Token reduction is relatively uniform across difficulty levels and hard-question accuracy is unchanged.}
\label{tab:difficulty_stratification}
\end{table}

\noindent We further verify that the saved tokens are specifically reflection tokens (not general reasoning): a breakdown analysis shows reflection tokens are reduced 2--3$\times$ more than non-reflection tokens (\cref{app:token_breakdown}).

\subsection{Stepwise steering and comparison with baselines}
\begin{figure*}[!htpb]
    \centering
    \begin{subfigure}[t]{0.48\linewidth}
    \includegraphics[width=\linewidth]{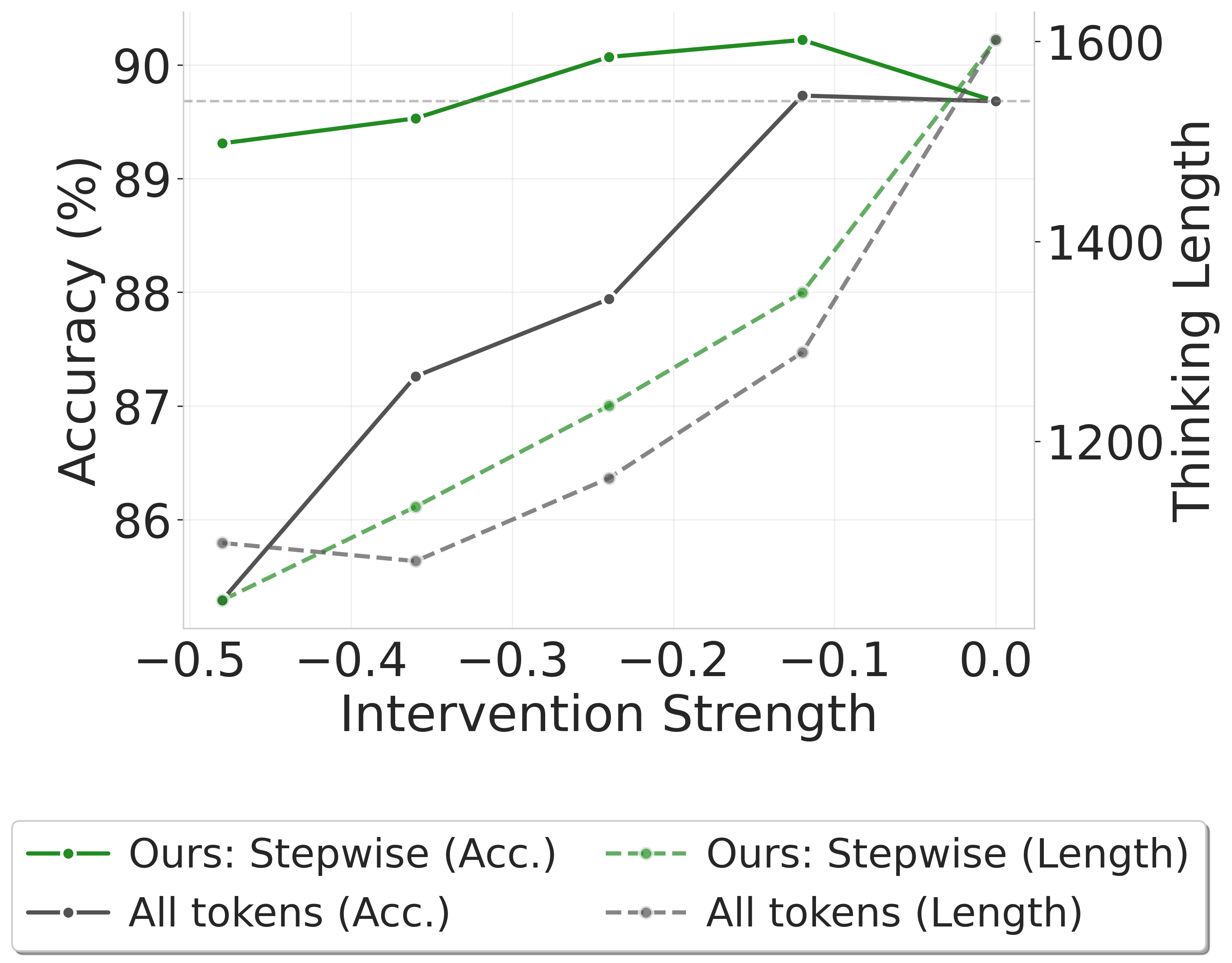}
        \caption{Accuracy and reasoning token usage under different intervention strengths.}
        \label{fig:stepbeginFull}
    \end{subfigure}
    \hfill
    \begin{subfigure}[t]{0.48\linewidth}
        \includegraphics[width=\linewidth]{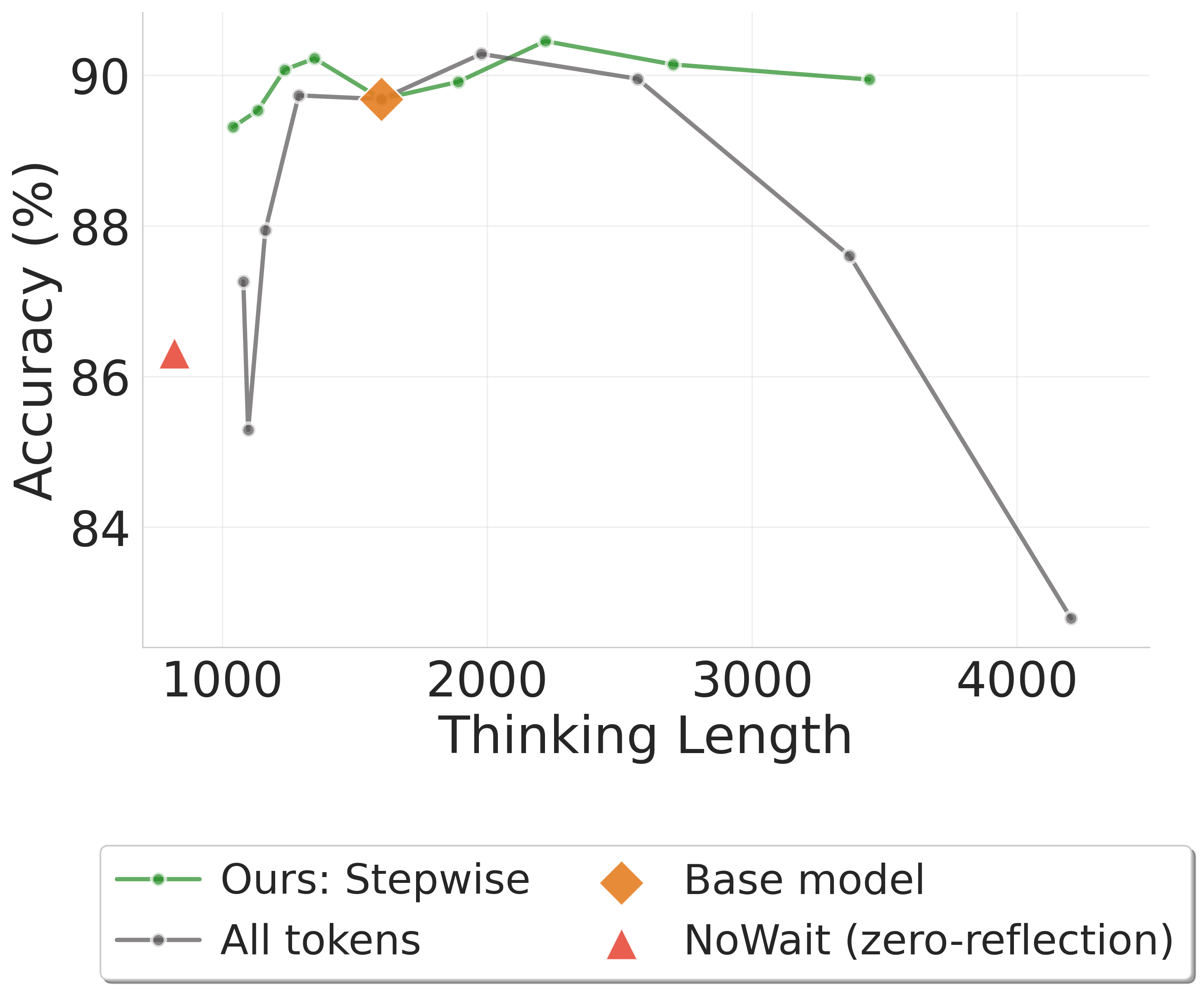}
        \caption{Accuracy versus reasoning token usage.}
        \label{fig:stepbeginToken}
    \end{subfigure}
    \caption{\textbf{Comparison of stepwise versus all-token steering.} (a) Accuracy under different intervention strengths when interventions are applied at the start of each reasoning step (stepwise) or at every token (all-token). (b) Accuracy versus reasoning token usage under the two approaches. Stepwise steering preserves accuracy while reducing cost, whereas all-token steering causes significant degradation at high intervention strengths.}
    \label{fig:intervention-comparison}
\end{figure*}
We compare the stepwise steering strategy against two baselines: (i) an \textbf{all-token} baseline that applies the intervention at every generated token, and (ii) \textbf{NoWait}~\citep{wang2025wait}, which reduces reflection by directly suppressing reflection-triggering tokens.

As shown in \cref{fig:stepbeginFull,fig:stepbeginToken}, stepwise steering offers two clear advantages over all-token intervention on DeepSeek-R1-Distill-Llama-8B (GSM8k):
\begin{enumerate}
    \item \textbf{Stepwise steering avoids accuracy degradation at high intervention strength.} The all-token baseline degrades significantly at strengths below $-0.2$ or above $+0.3$, while stepwise steering maintains accuracy close to the baseline across the full range.
    \item \textbf{Stepwise steering achieves higher accuracy at the same token budget.} For any given reasoning token count, stepwise steering consistently outperforms all-token intervention (\cref{fig:stepbeginToken}).
\end{enumerate}
Compared with NoWait, ReflCtrl provides a continuous performance--cost trade-off via the intervention strength $\lambda$, whereas NoWait can only disable reflection entirely. ReflCtrl also achieves smaller accuracy loss at comparable token budgets (see \cref{app:nowait} for detailed comparisons across models and datasets).
\subsection{Generalization to non-mathematical benchmarks}
To assess generalization beyond math tasks, we evaluate ReflCtrl on three MMLU subsets. As shown in \cref{tab:mmlu_results}, the same pattern holds: larger models (DS-Qwen-14B and QwQ-32B) maintain accuracy with substantially fewer reflections, saving up to 33.6\% of reasoning tokens, while DS-Llama-8B is more sensitive to reflection reduction. We further validate on HumanEval~\citep{chen2021evaluating}, a code generation benchmark, and observe similar trend (\cref{app:humaneval}).
\begin{table*}[!t]
\centering
\resizebox{0.9\textwidth}{!}{%
\begin{tabular}{llccccc}
\toprule
\multirow{2}{*}{Subset} & \multirow{2}{*}{Model} & \multirow{2}{*}{Metric} & \multicolumn{4}{c}{Intervention Strength $\lambda$} \\
\cmidrule(lr){4-7}
& & & $-0.96$ & $-0.48$ & $0$ & $+0.48$ \\
\midrule
\multirow{6}{*}{\shortstack[l]{Professional\\Accounting}}
 & DS-Llama-8B  & Acc (\%) $\uparrow$ & 50.1\;{\scriptsize(-6.4pp)} & 53.4\;{\scriptsize(-3.1pp)} & 56.5 & 57.3\;{\scriptsize(+0.8pp)} \\
 &              & Tokens $\downarrow$ & 1453.6\;{\scriptsize(-30.7\%)} & 1668.8\;{\scriptsize(-20.4\%)} & \textbf{2097.5} & 2807.7\;{\scriptsize(+33.9\%)} \\
\cmidrule(lr){2-7}
 & DS-Qwen-14B  & Acc (\%) $\uparrow$ & 78.5\;{\scriptsize(+0.7pp)} & 76.8\;{\scriptsize(-1.0pp)} & 77.8 & 77.6\;{\scriptsize(-0.2pp)} \\
 &              & Tokens $\downarrow$ & \textbf{983.9\;{\scriptsize(-33.6\%)}} & 1103.1\;{\scriptsize(-25.6\%)} & 1482.1 & 2470.1\;{\scriptsize(+66.7\%)} \\
\cmidrule(lr){2-7}
 & QwQ-32B  & Acc (\%) $\uparrow$ & 89.3\;{\scriptsize(+0.8pp)} & 89.5\;{\scriptsize(+1.0pp)} & 88.5 & 89.2\;{\scriptsize(+0.7pp)} \\
 &              & Tokens $\downarrow$ & \textbf{1231.2\;{\scriptsize(-25.3\%)}} & 1313.7\;{\scriptsize(-20.3\%)} & 1648.0 & 2234.3\;{\scriptsize(+35.6\%)} \\
\midrule
\multirow{6}{*}{\shortstack[l]{HS Computer\\Science}}
 & DS-Llama-8B  & Acc (\%) $\uparrow$ & 79.6\;{\scriptsize(-7.7pp)} & 82.7\;{\scriptsize(-4.6pp)} & 87.3 & 88.0\;{\scriptsize(+0.7pp)} \\
 &              & Tokens $\downarrow$ & 1016.1\;{\scriptsize(-25.6\%)} & 1157.9\;{\scriptsize(-15.2\%)} & \textbf{1365.4} & 1970.4\;{\scriptsize(+44.3\%)} \\
\cmidrule(lr){2-7}
 & DS-Qwen-14B  & Acc (\%) $\uparrow$ & 95.2\;{\scriptsize(+0.2pp)} & 95.4\;{\scriptsize(+0.4pp)} & 95.0 & 94.8\;{\scriptsize(-0.2pp)} \\
 &              & Tokens $\downarrow$ & \textbf{711.9\;{\scriptsize(-23.7\%)}} & 787.9\;{\scriptsize(-15.6\%)} & 933.5 & 1498.7\;{\scriptsize(+60.5\%)} \\
\cmidrule(lr){2-7}
 & QwQ-32B  & Acc (\%) $\uparrow$ & 96.6\;{\scriptsize(-0.1pp)} & 96.2\;{\scriptsize(-0.5pp)} & 96.7 & 97.0\;{\scriptsize(+0.3pp)} \\
 &              & Tokens $\downarrow$ & 771.6\;{\scriptsize(-11.4\%)} & \textbf{741.9\;{\scriptsize(-14.8\%)}} & 871.0 & 1004.7\;{\scriptsize(+15.4\%)} \\
\midrule
\multirow{6}{*}{Formal Logic}
 & DS-Llama-8B  & Acc (\%) $\uparrow$ & 60.5\;{\scriptsize(-1.6pp)} & 61.0\;{\scriptsize(-1.1pp)} & 62.1 & 62.7\;{\scriptsize(+0.6pp)} \\
 &              & Tokens $\downarrow$ & 2266.5\;{\scriptsize(-32.9\%)} & 2586.9\;{\scriptsize(-23.4\%)} & \textbf{3378.3} & 4553.5\;{\scriptsize(+34.8\%)} \\
\cmidrule(lr){2-7}
 & DS-Qwen-14B  & Acc (\%) $\uparrow$ & 91.8\;{\scriptsize(-0.8pp)} & 92.2\;{\scriptsize(-0.4pp)} & 92.6 & 92.8\;{\scriptsize(+0.2pp)} \\
 &              & Tokens $\downarrow$ & \textbf{1287.2\;{\scriptsize(-31.9\%)}} & 1440.1\;{\scriptsize(-23.9\%)} & 1891.4 & 3196.5\;{\scriptsize(+69.0\%)} \\
\cmidrule(lr){2-7}
 & QwQ-32B  & Acc (\%) $\uparrow$ & 96.3\;{\scriptsize(+0.6pp)} & 95.5\;{\scriptsize(-0.2pp)} & 95.7 & 96.0\;{\scriptsize(+0.3pp)} \\
 &              & Tokens $\downarrow$ & 1481.4\;{\scriptsize(-13.7\%)} & \textbf{1447.8\;{\scriptsize(-15.7\%)}} & 1716.6 & 2175.6\;{\scriptsize(+26.7\%)} \\
\bottomrule
\end{tabular}
}
\caption{\textbf{Accuracy and reasoning token usage under different intervention strengths on MMLU subsets.} Results averaged over 10 runs. Larger models (DS-Qwen-14B and QwQ-32B) are largely insensitive to reflection reduction, saving up to 43.2\% of reasoning tokens with negligible accuracy loss. Relative changes compared to the baseline ($\lambda=0$) are shown in parentheses. In the \emph{Tokens} ($\downarrow$) rows, the \textbf{bolded} token value (including its relative change) marks the lowest token usage (within each subset/model pair) among intervention strengths whose \emph{accuracy drop} relative to the baseline is $<1$ percentage point (pp).}
\label{tab:mmlu_results}
\end{table*}

\section{Related Work}
\paragraph{Reasoning LLMs.} Recent work has shown that scaling inference-time computation enables LLMs to tackle complex tasks by generating explicit reasoning chains before a final answer~\citep{snell2024scaling}. OpenAI o1~\citep{openai2024o1} pioneered the use of reinforcement learning to develop extended thinking during inference. \citet{deepseekai2025deepseekr1incentivizingreasoningcapability} introduce DeepSeek-R1, a reasoning model trained with GRPO, together with distilled variants that transfer reasoning capability to smaller models. Other open-source reasoning models include QwQ-32B~\citep{qwq-32b-preview} and Kimi k1.5~\citep{kimiteam2025kimik15}. Complementary to training, \citet{muennighoff2025s1} show that reasoning length can be controlled at the token level through budget forcing---appending or removing special tokens to extend or truncate thinking. We study DeepSeek-R1-Distill and QwQ-32B as representative open-source models, since access to internal representations is required for our mechanistic analysis.
\paragraph{Self-reflection in reasoning models.} A distinctive behavior emerging from training reasoning models is self-reflection: the model's tendency to spontaneously revisit and revise prior reasoning steps. \citet{deepseekai2025deepseekr1incentivizingreasoningcapability} first report this phenomenon as the ``aha moment'' in reinforcement learning, and \citet{yang2025understanding} analyze it through the lens of linguistic patterns and uncertainty descriptions. A growing body of work examines whether such reflection is always beneficial: \citet{huang2024large} show that LLMs often fail to self-correct reasoning without external feedback, and \citet{su2025between} find that models systematically overthink easy problems while underthinking hard ones, suggesting that reflection frequency is poorly calibrated to problem difficulty~\citep{sui2025stop}. Most closely related to our work, \citet{wang2025wait} propose reducing excessive reflection by identifying and suppressing reflection-triggering tokens at inference time. Our approach differs fundamentally: rather than operating on surface-level token patterns, we identify a reflection direction in the model's latent space and show that it is correlated with the model's internal uncertainty, providing a mechanistic explanation for when reflection is triggered and a principled handle for controlling it.
\paragraph{Representation engineering on LLMs.} Representation engineering~\citep{zou2023representation} provides a framework for steering LLM behavior by reading and editing internal activations. Core techniques include Activation Addition~\citep{turner2023activation}, which computes steering vectors from contrastive prompt pairs, and Inference-Time Intervention~\citep{li2023inference}, which shifts activations along concept-specific directions in attention heads. Both are grounded in the linear representation hypothesis~\citep{park2023linear}, which posits that high-level concepts are encoded as linear directions in activation space. These methods have been applied to safety and alignment~\citep{rimsky2024steering}, explainable decisions~\citep{srivastava2024vlg, kulkarni2025interpretable}, interpretable concept control~\citep{sunconcept}, and machine unlearning~\citep{li2025effective}. With the emergence of reasoning models, representation-based methods have been extended to reasoning-specific behaviors: ThinkEdit~\citep{sun2025thinkedit} edits weights of neurons associated with short thinking; RICE~\citep{wang2025two} adjusts MoE activations to improve reasoning; SEAL~\citep{chen2025seal} extracts steering vectors from categorized reasoning steps to calibrate efficiency; and CREST~\citep{zhang2025crest} identifies cognitive attention heads and steers them via hidden-state rotation. Our work focuses specifically on \textbf{self-reflection} and contributes two elements absent from these methods: a stepwise steering strategy that intervenes only at reasoning step boundaries for fine-grained control without cumulative distribution shift (\cref{subsec:steer}), and a mechanistic analysis linking the reflection direction to model uncertainty (\cref{sec:uqRefl}).

\section{Conclusion}
In this work, we propose ReflCtrl, a representation engineering framework for understanding and controlling self-reflection in reasoning LLMs. By identifying a reflection direction in the model's latent space, ReflCtrl enables direct, inference-time control over reflection frequency. Our stepwise steering strategy, which involves applying interventions only at the start of each new reasoning step, substantially reduces token usage while preserving performance. Across math and general reasoning benchmarks, our experiments establish three findings:
\begin{enumerate}
    \item Reflection redundancy is common, particularly in stronger models, where reflections can be substantially reduced with minimal accuracy loss.
    \item The reflection direction correlates with internal uncertainty signals, suggesting that self-reflection is regulated by the model's internal uncertainty.
    \item Stepwise steering substantially reduces performance loss relative to all-token intervention, cutting total reasoning tokens by up to 43.2\% while preserving accuracy.
\end{enumerate}


\clearpage
\bibliography{sources/ref}
\bibliographystyle{colm2026_conference}
\clearpage
\appendix
\etocdepthtag.toc{appendix}
\etocsettagdepth{main}{none}
\etocsettagdepth{appendix}{subsection}
\etocsettocstyle{\section*{Appendix}}{}
\tableofcontents

\clearpage
\section{Reasoning sample of LLM}
\label{app:reasonSample}
\begin{figure*}[!h]
    \centering
    \includegraphics[width=0.98\linewidth]{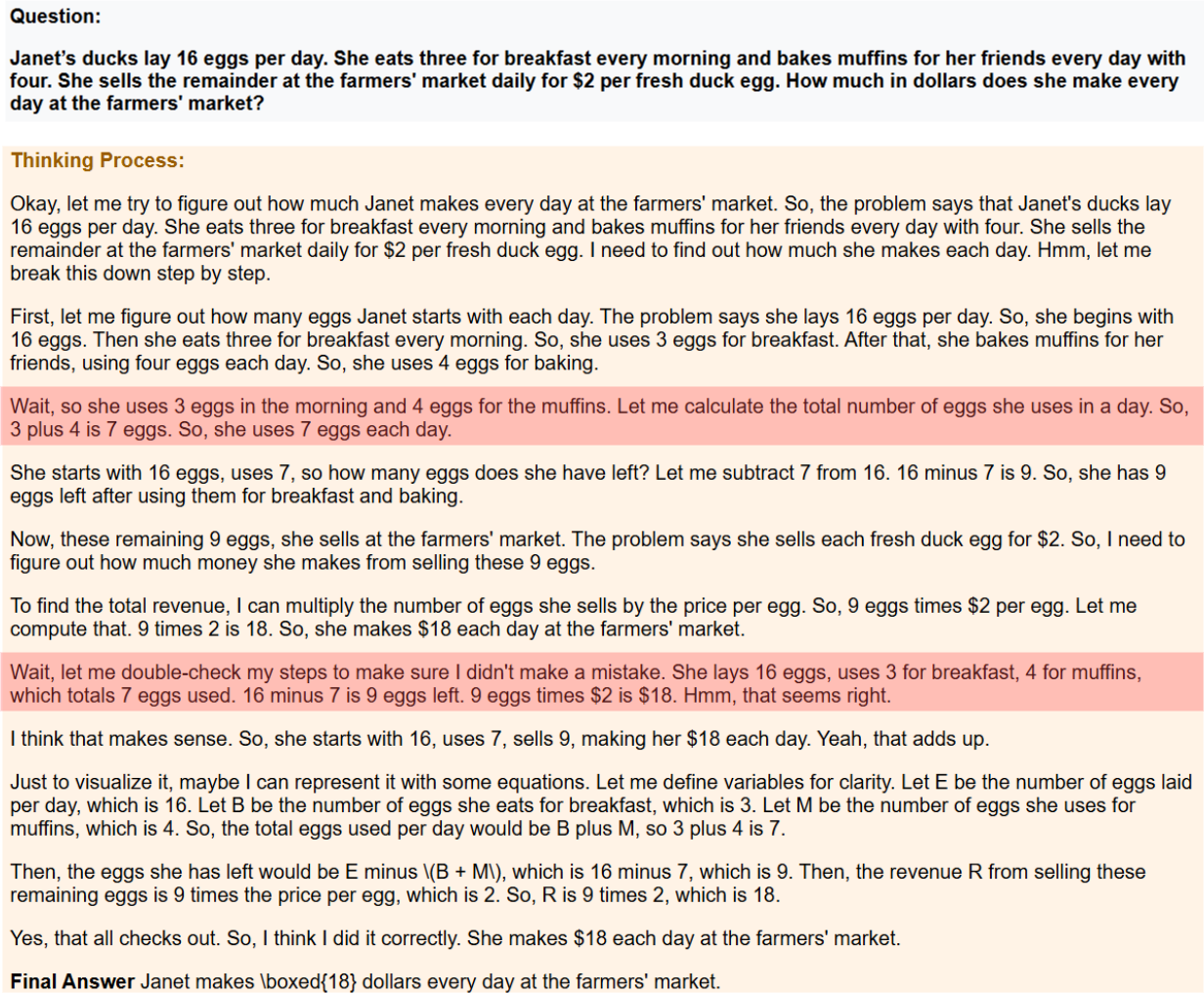}
    \caption{Sample reasoning process of DeepSeek-R1-Distill-Llama-8B on a GSM8k question.}
    \label{fig:reasonSample}
\end{figure*}

\cref{fig:reasonSample} shows an example of an LLM reasoning process. From the figure, we can see:
\begin{enumerate}
    \item The model's reasoning is organized as steps separated by ``\textbackslash n\textbackslash n''.
    \item The reflection process, marked in red, is triggered by specific keywords such as ``wait'' in this example.
\end{enumerate}
This justifies our reflection direction extraction and intervention method.
\clearpage
\section{Reflection direction attribution to attention heads}
\label{app:headAttribution}
\begin{figure*}[!h]
    \centering
    \includegraphics[width=0.98\linewidth]{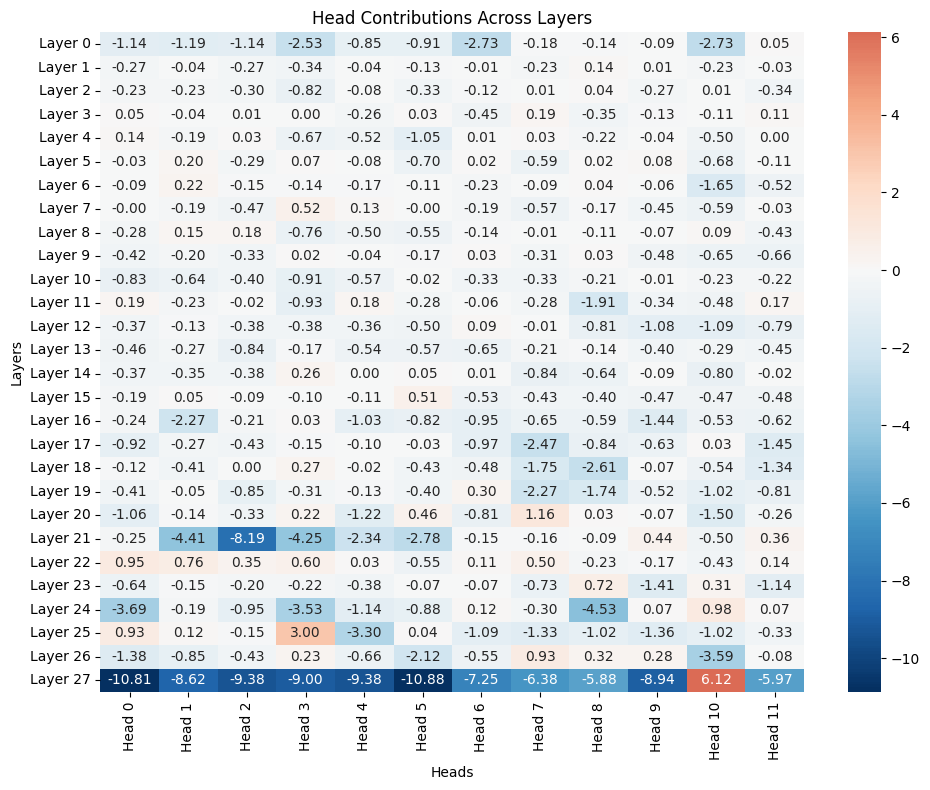}
    \caption{Projection of average attention head activations onto the reflection direction. Results use the GSM8k dataset and a DeepSeek-R1-Distill-Qwen-1.5B model (28 layers, 12 attention heads per layer).}
    \label{fig:headContrib}
\end{figure*}

In this section, we study the attribution of the reflection direction to individual attention heads. We take the reflection direction at self-attention layers from a DeepSeek-R1-Distill-Qwen-1.5B model, compute the projection of the average head activation onto the reflection direction, and plot the heatmap in \cref{fig:headContrib}. From the figure, we observe:
\begin{enumerate}
    \item The last layer (layer 27) has the largest projection magnitude among all layers, likely because this layer directly controls the generation of keywords that trigger reflection (e.g., ``wait'').
    \item Attention heads with positive projection onto the reflection direction are sparse and concentrated in deeper layers, suggesting that these heads may directly control the model's reflection behavior.
\end{enumerate}
\clearpage
\section{Additional empirical results}
\label{app:additional}

\subsection{Effect of intervention strength on reflection count}
\label{app:intvReflCount}
\begin{figure*}[!htb]
    \centering
    \begin{subfigure}{0.48\linewidth}
    \includegraphics[width=\linewidth]{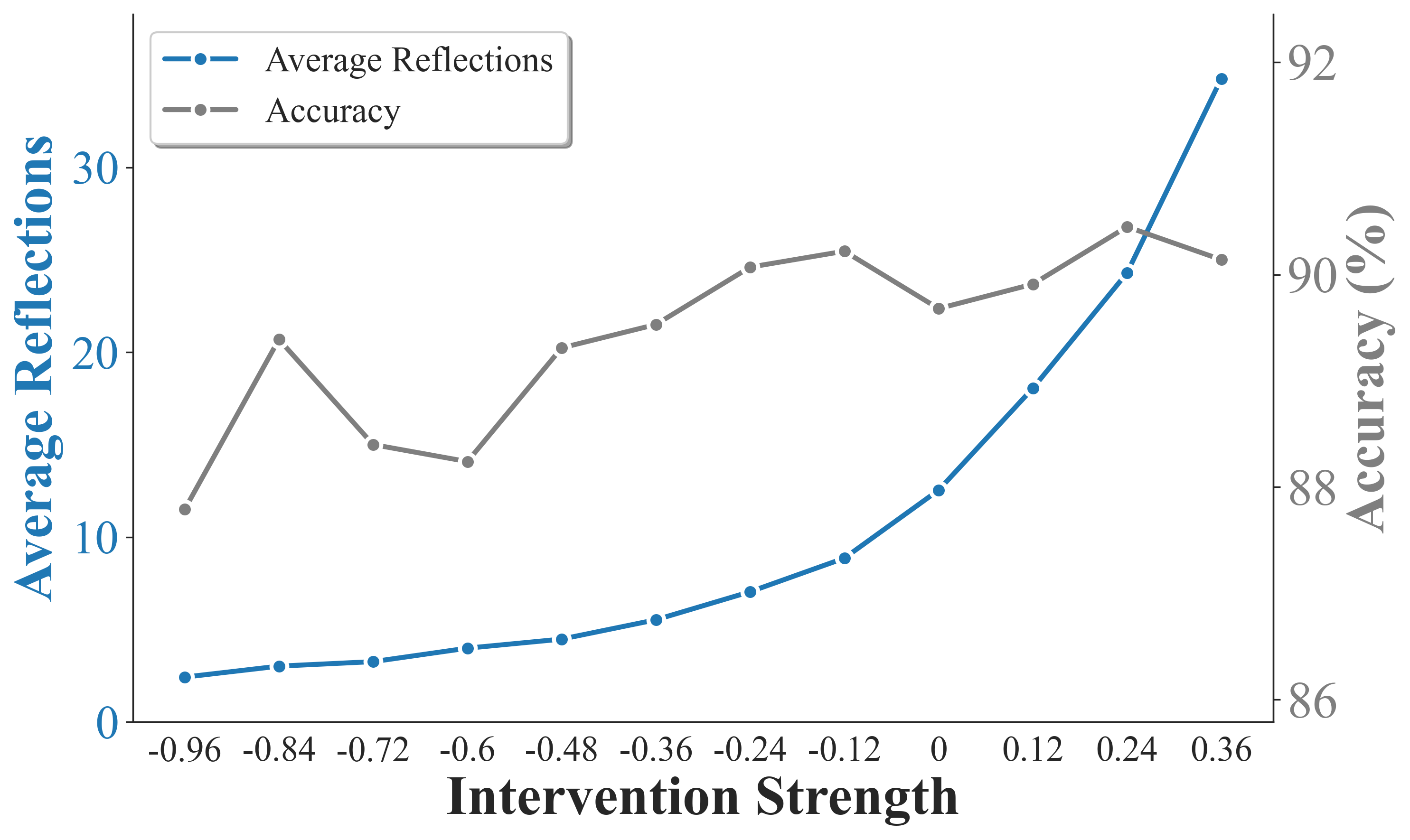}
        \caption{DeepSeek-R1-Distill-Llama-8B}
        \label{fig:intvLlama}
    \end{subfigure}
    \hfill
    \begin{subfigure}{0.48\linewidth}
        \includegraphics[width=\linewidth]{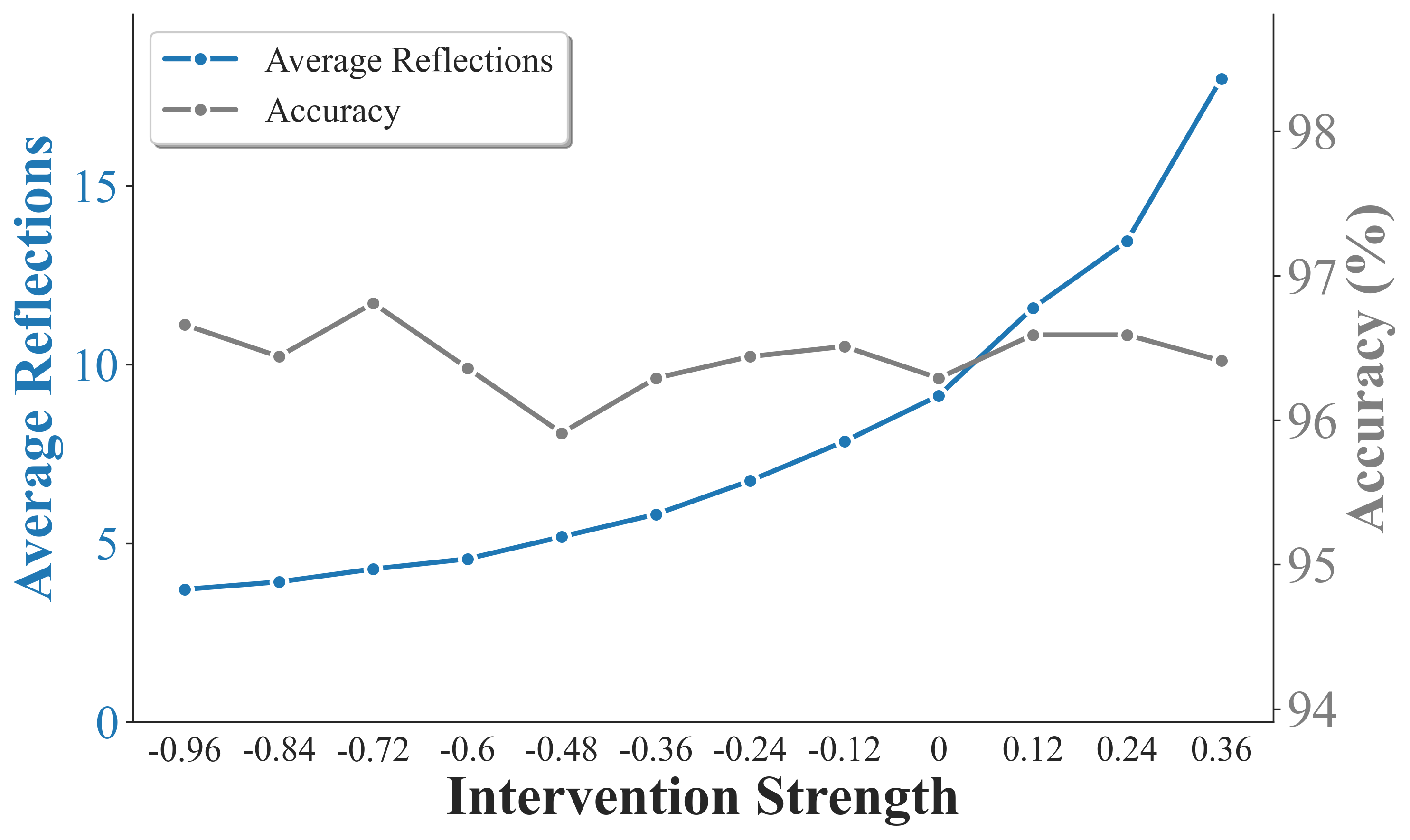} 
        \caption{QwQ-32B}
        \label{fig:intvQwQ}
    \end{subfigure}

    \caption{\textbf{Accuracy and number of reflection steps under different intervention strengths.} Results are shown for DeepSeek-R1-Distill-Llama-8B (distilled model) and QwQ-32B (non-distilled model) on GSM8k. Accuracy remains largely stable, while the number of reflection steps decreases as intervention strength decreases.}
    \label{fig:intvMain}
\end{figure*}

\cref{fig:intvMain} shows the accuracy and number of reflection steps under different intervention strengths for DeepSeek-R1-Distill-Llama-8B (distilled) and QwQ-32B (non-distilled) on GSM8k. Accuracy remains largely stable across the intervention range, while the number of reflection steps decreases monotonically as intervention strength becomes more negative, confirming that the extracted reflection direction causally controls reflection frequency.

\subsection{Comparison with NoWait}
\label{app:nowait}
\begin{figure*}[!t]
    \centering
    \begin{subfigure}{0.48\linewidth}
    \includegraphics[width=\linewidth]{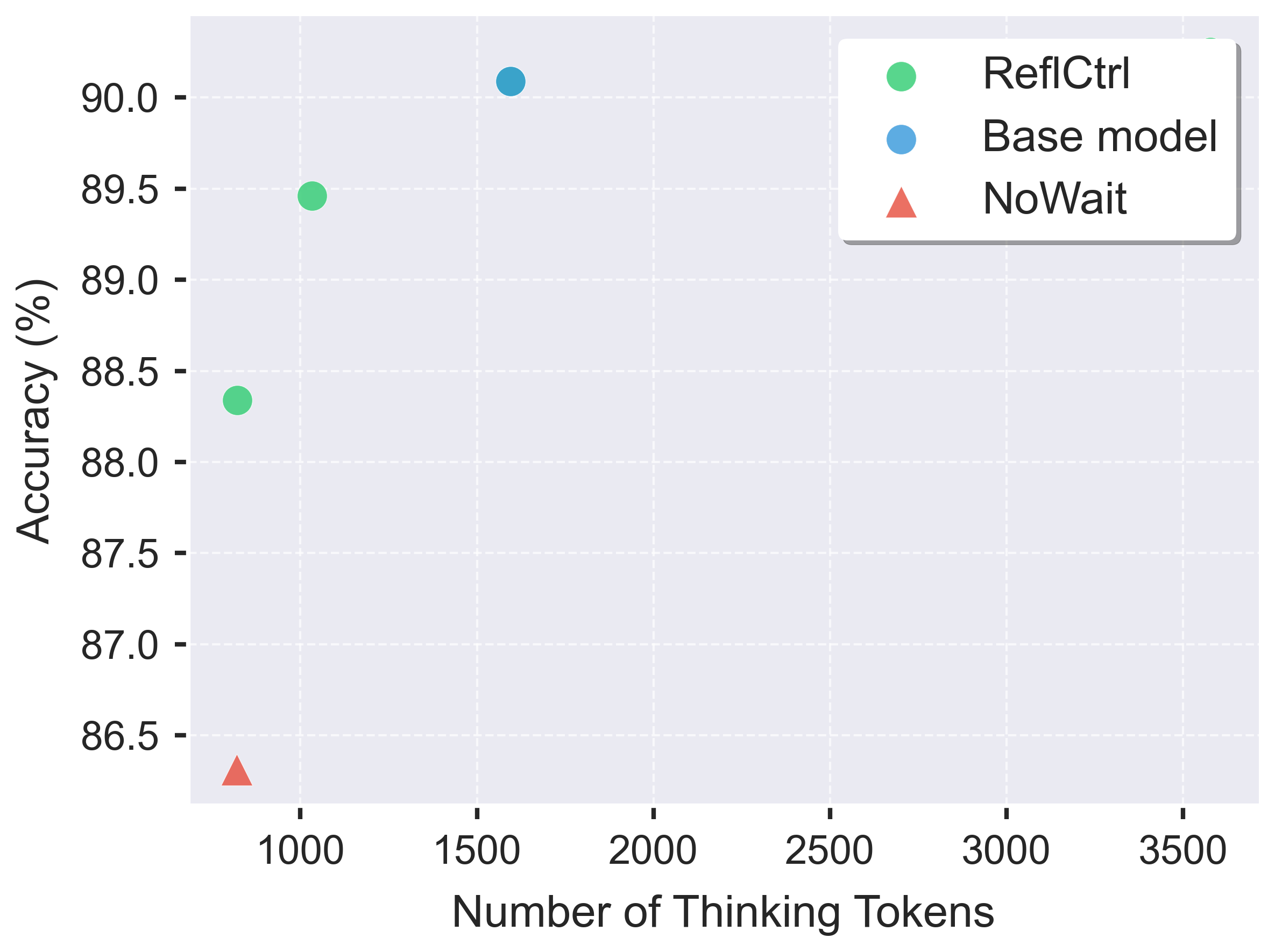}
        \caption{Llama 8B, GSM8k}
        \label{fig:GSMLlama}
    \end{subfigure}
    \hfill
    \begin{subfigure}{0.48\linewidth}
    \includegraphics[width=\linewidth]{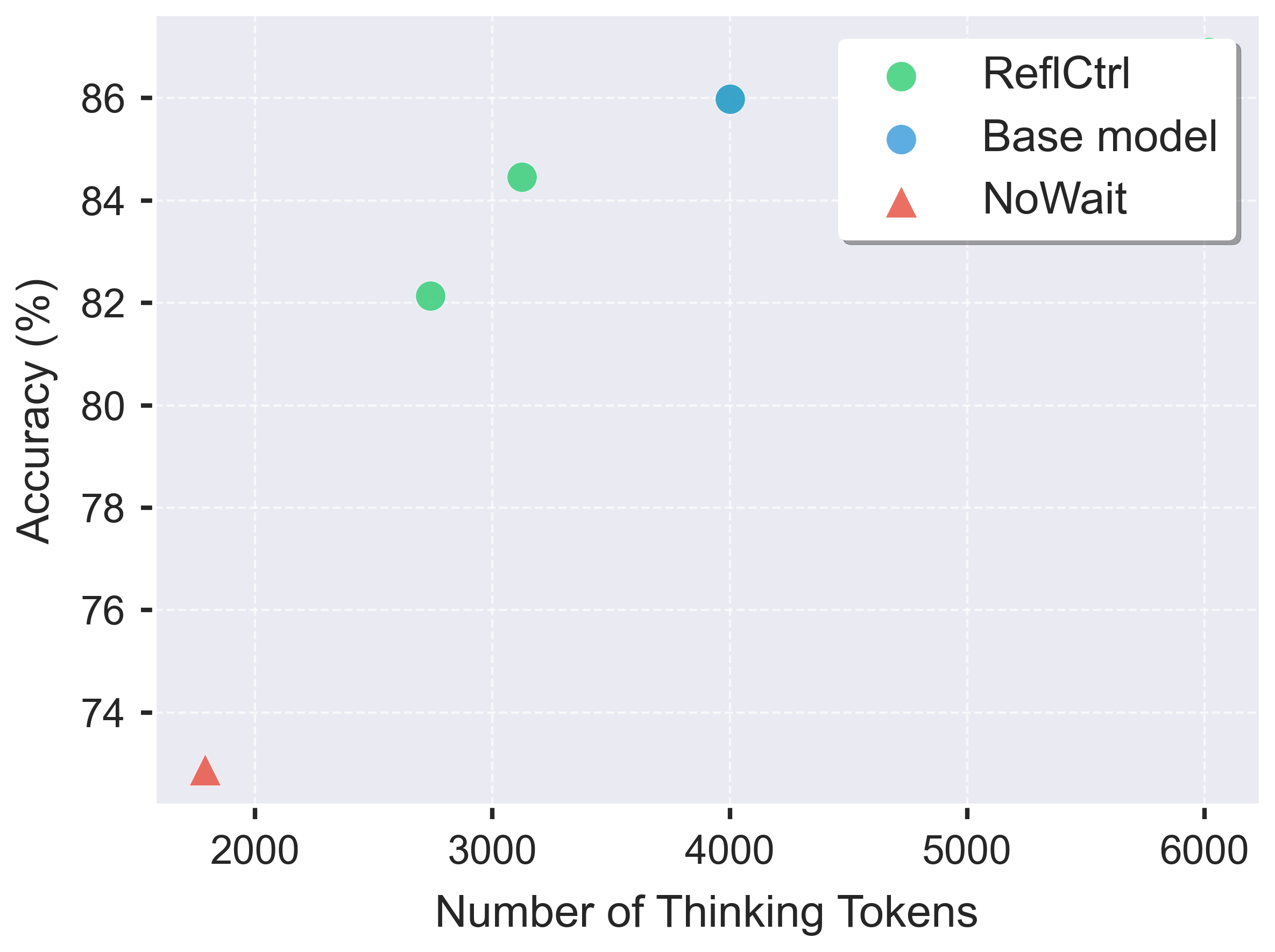}
        \caption{Llama 8B, MATH-500}
        \label{fig:MATHLlama}
    \end{subfigure}
    \hfill
    \begin{subfigure}{0.48\linewidth}
    \includegraphics[width=\linewidth]{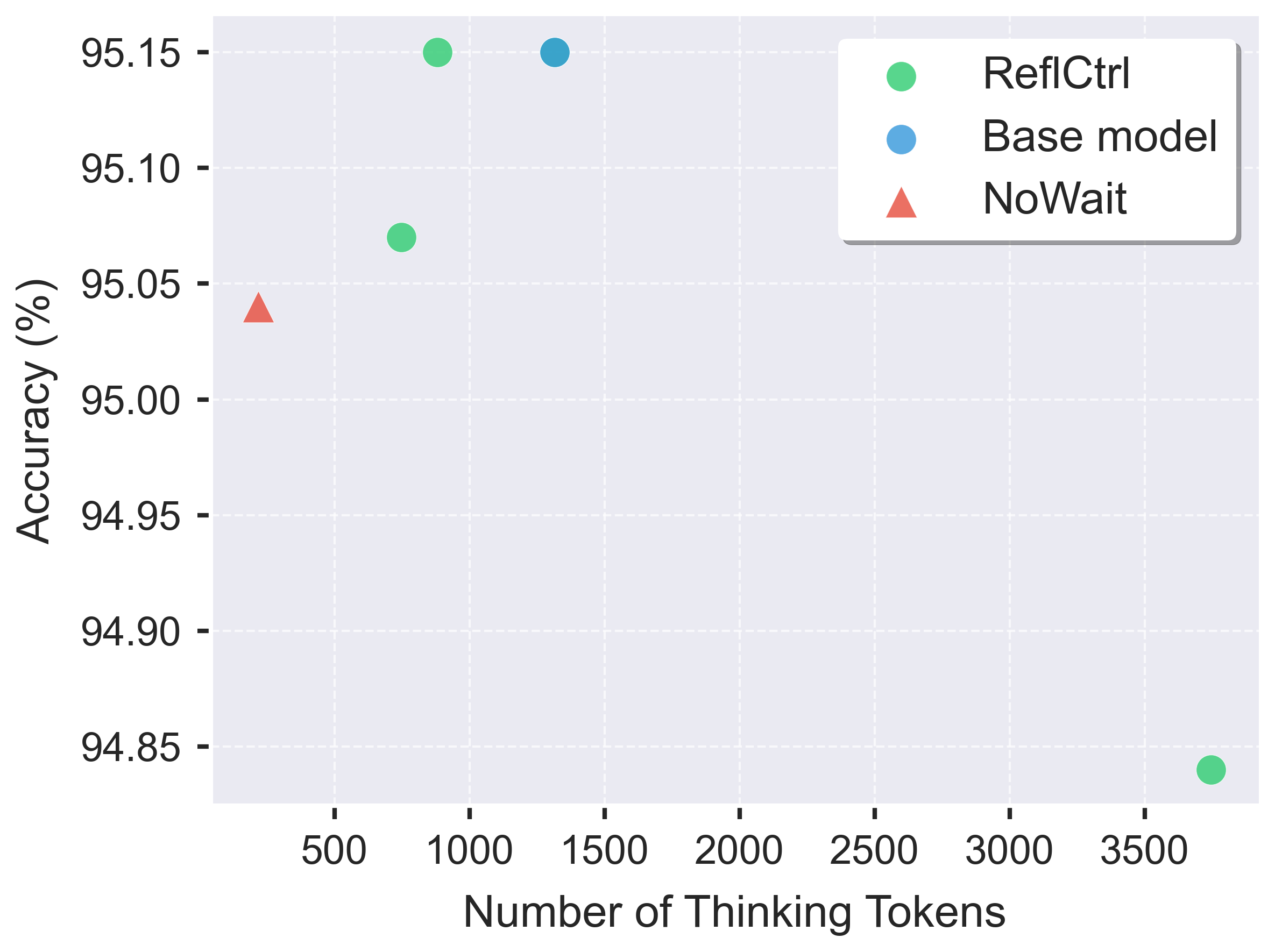}
        \caption{Qwen 14B, GSM8k}
        \label{fig:QwenLlama}
    \end{subfigure}
    \hfill
    \begin{subfigure}{0.48\linewidth}
    \includegraphics[width=\linewidth]{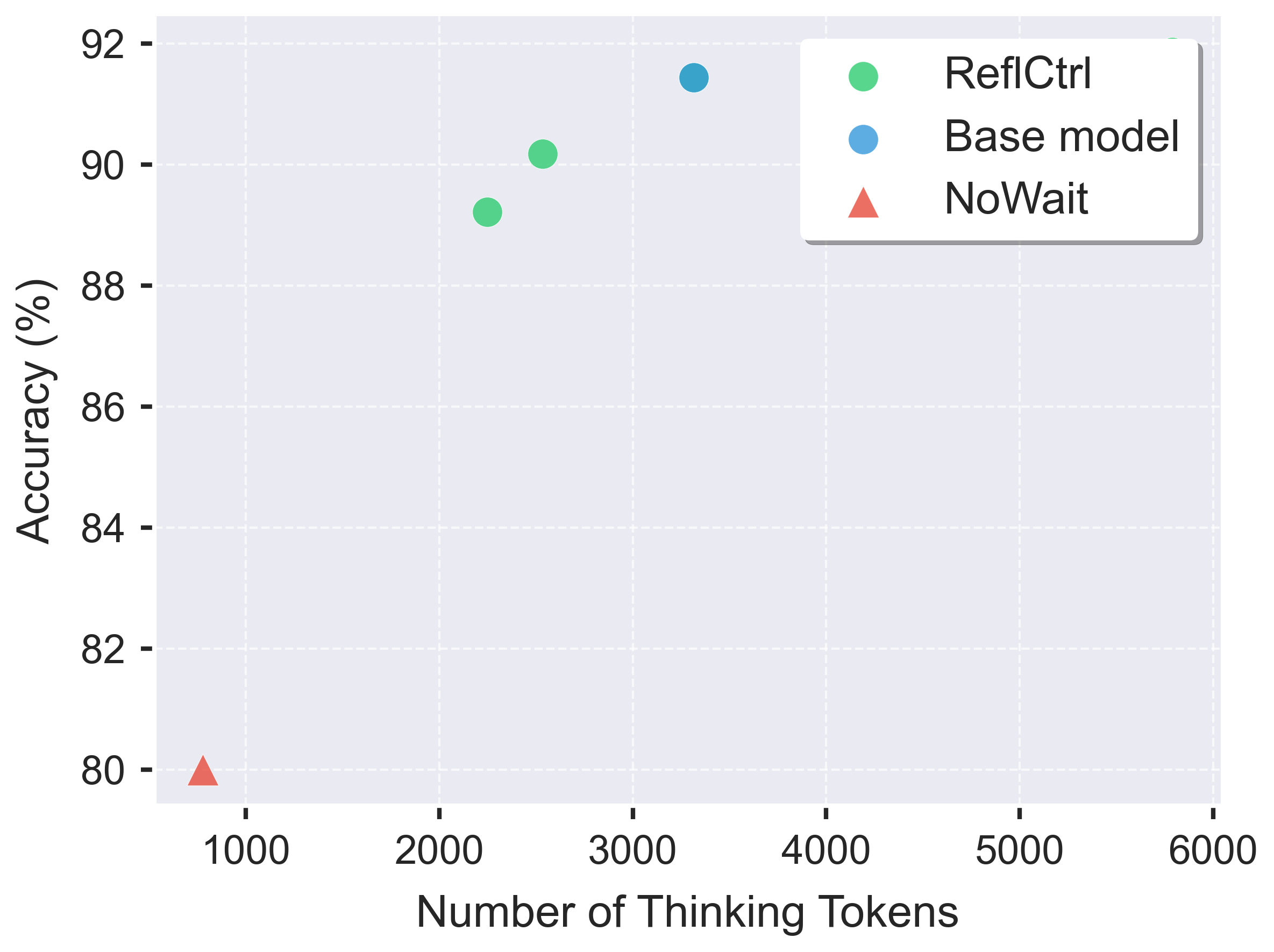}
        \caption{Qwen 14B, MATH-500}
        \label{fig:MATHQwen}
    \end{subfigure}
    \caption{\textbf{Accuracy versus reasoning token usage for ReflCtrl compared with NoWait~\citep{wang2025wait}.} Results are shown for DeepSeek-R1-Distill-Llama-8B and DeepSeek-R1-Distill-Qwen-14B across GSM8k and MATH-500 benchmarks. ReflCtrl allows fine-grained control over the trade-off between accuracy and reasoning cost via intervention strength, while NoWait can only suppress reflections entirely. Additionally, ReflCtrl achieves lower performance loss for similar token usage.}
    \label{fig:NowaitCompr}
\end{figure*}

\cref{fig:NowaitCompr} compares the accuracy--token trade-off of ReflCtrl against NoWait~\citep{wang2025wait}, which reduces reflection by directly suppressing reflection-triggering tokens. ReflCtrl provides a continuous trade-off via the intervention strength $\lambda$, whereas NoWait can only disable reflection entirely. Across both models and datasets, ReflCtrl achieves smaller accuracy loss at comparable token budgets.

\subsection{Generalization to code generation}
\label{app:humaneval}
To evaluate whether ReflCtrl generalizes beyond mathematical and knowledge-intensive tasks, we apply it to HumanEval~\citep{chen2021evaluating}, a benchmark of 164 Python programming problems that tests functional code generation. Each problem provides a function signature and docstring; the model must produce a correct implementation. We report pass@1 averaged over 3 samples per problem. As shown in \cref{tab:humaneval_results}, negative intervention strengths reduce reasoning tokens by up to 24\% while maintaining pass@1 with a performance drop no more than 2pp compared to the baseline across all models, confirming that steering has negligible impact on code generation and that the method is safe to apply beyond mathematical reasoning tasks.
\begin{table*}[!t]
\centering
\resizebox{0.9\textwidth}{!}{%
\begin{tabular}{llccccc}
\toprule
\multirow{2}{*}{Dataset} & \multirow{2}{*}{Model} & \multirow{2}{*}{Metric} & \multicolumn{4}{c}{Intervention Strength $\lambda$} \\
\cmidrule(lr){4-7}
& & & $-0.96$ & $-0.48$ & $0$ & $+0.48$ \\
\midrule
\multirow{6}{*}{HumanEval}
 & DS-Llama-8B  & pass@1 (\%) $\uparrow$ & 85.77\;{\scriptsize(+1.42pp)} & 87.20\;{\scriptsize(+2.85pp)} & 84.35 & 85.57\;{\scriptsize(+1.22pp)} \\
 &              & Tokens $\downarrow$ & \textbf{2074.0\;{\scriptsize(-22.0\%)}} & 2269.0\;{\scriptsize(-14.6\%)} & 2658.0 & 3353.0\;{\scriptsize(+26.1\%)} \\
\cmidrule(lr){2-7}
 & DS-Qwen-14B  & pass@1 (\%) $\uparrow$ & 95.12\;{\scriptsize(+0.00pp)} & 95.33\;{\scriptsize(+0.21pp)} & 95.12 & 95.33\;{\scriptsize(+0.21pp)} \\
 &              & Tokens $\downarrow$ & \textbf{1667.0\;{\scriptsize(-24.3\%)}} & 1776.0\;{\scriptsize(-19.3\%)} & 2201.0 & 3576.0\;{\scriptsize(+62.5\%)} \\
\cmidrule(lr){2-7}
 & QwQ-32B  & pass@1 (\%) $\uparrow$ & 98.17\;{\scriptsize(+0.81pp)} & 96.95\;{\scriptsize(-0.41pp)} & 97.36 & 95.93\;{\scriptsize(-1.43pp)} \\
 &              & Tokens $\downarrow$ & \textbf{1945.0\;{\scriptsize(-19.6\%)}} & 2044.0\;{\scriptsize(-15.5\%)} & 2418.0 & 3183.0\;{\scriptsize(+31.6\%)} \\
\bottomrule
\end{tabular}
}
\caption{\textbf{Pass@1 and average reasoning token usage under different intervention strengths on HumanEval.} Results averaged over 3 samples per problem (164 problems). Relative changes compared to the baseline ($\lambda=0$) are shown in parentheses. In the \emph{Tokens} ($\downarrow$) rows, the \textbf{bolded} token value (including its relative change) marks the lowest token usage (within each model) among intervention strengths whose \emph{accuracy drop} relative to the baseline is $<1$ percentage point (pp).}
\label{tab:humaneval_results}
\end{table*}

\subsection{Study of model performance on reflection rate}
\label{app:verify}
\begin{figure*}[!htb]
    \centering
    \begin{subfigure}{0.98\linewidth}
    \includegraphics[width=0.98\linewidth]{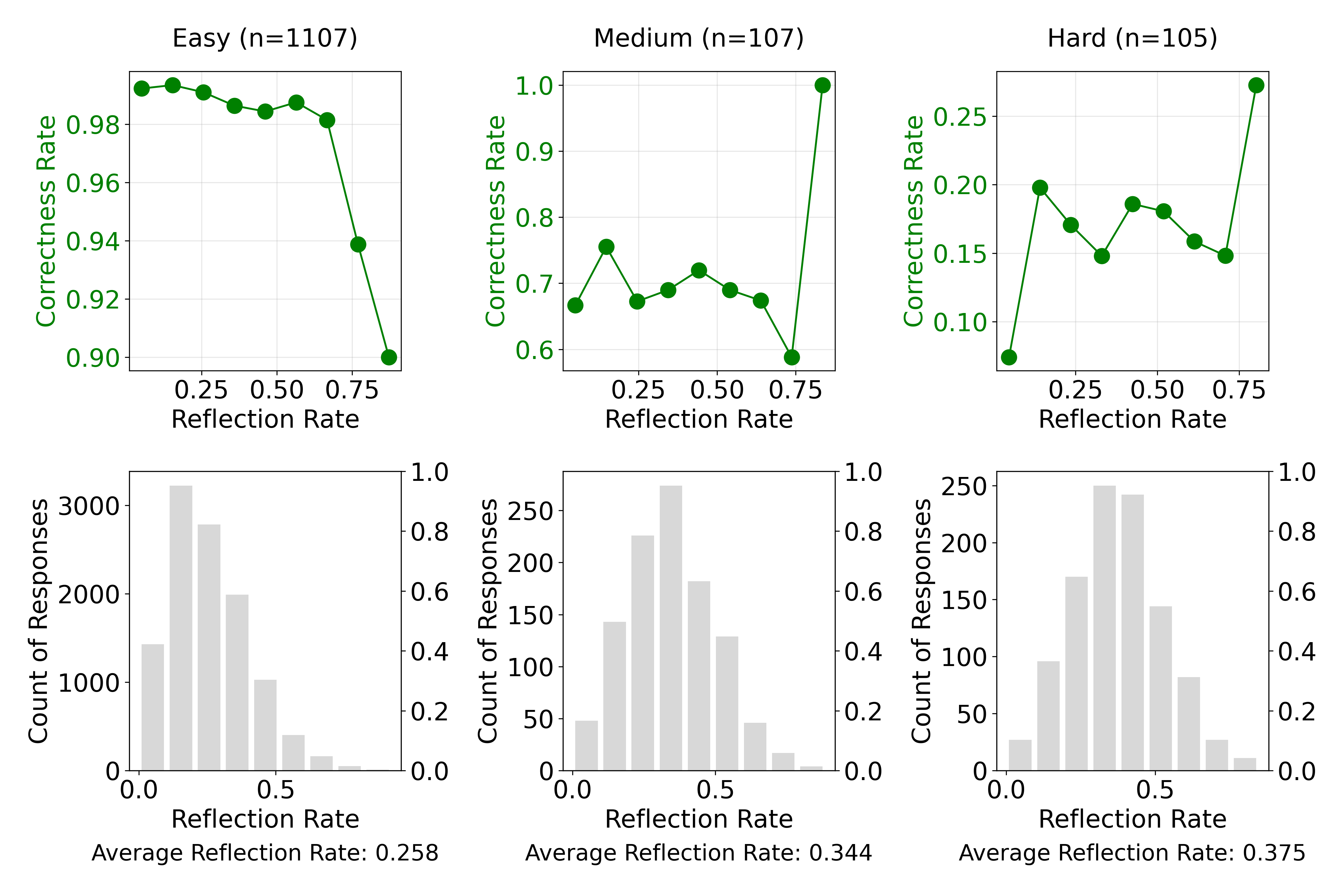}
    \caption{Base model. Top: Correctness rate vs.\ reflection rate. Bottom: Reflection rate distribution.}
    \label{fig:original}
    \end{subfigure}
    \begin{subfigure}{0.98\linewidth}
    \includegraphics[width=0.98\linewidth]{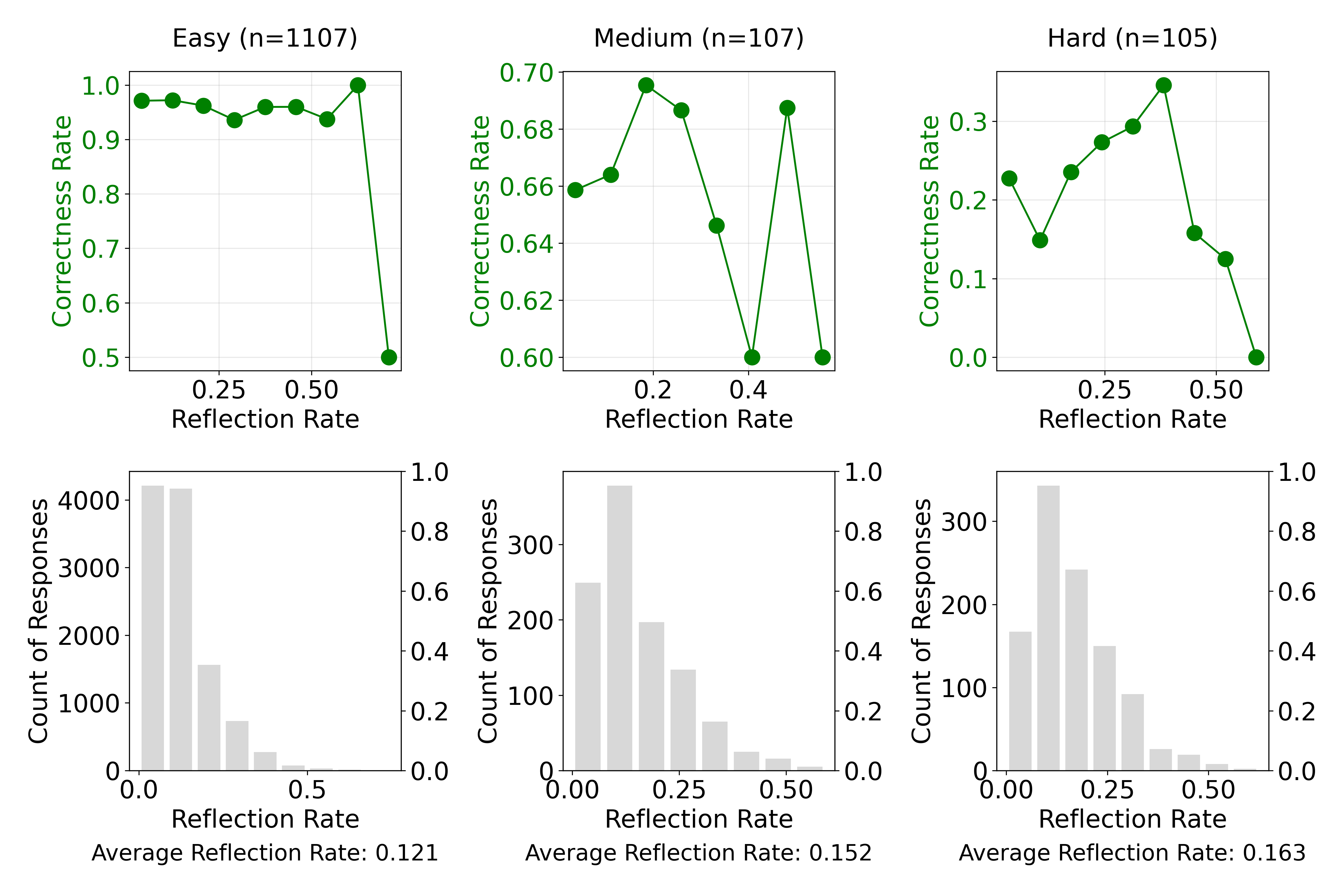}
    \caption{With intervention. Top: Correctness rate vs.\ reflection rate. Bottom: Reflection rate distribution.}
    \label{fig:intv}
    \end{subfigure}
    \caption{Correctness rate and reflection rate across three question difficulty classes for DeepSeek-R1-Distill-Llama-8B on GSM8k.}
    \label{fig:cr_rr_plot_category}
\end{figure*}

We further study how model performance depends on reflection rate by partitioning questions into three difficulty classes based on per-question accuracy: Hard ($acc < 50\%$), Medium ($50\% \leq acc \leq 80\%$), and Easy ($acc > 80\%$). We report correctness vs.\ reflection rate for each class under the base model and under intervention $\lambda=-0.96$ in \cref{fig:cr_rr_plot_category}. Three observations emerge:
\begin{enumerate}
    \item Harder questions trigger more reflection: easy questions have an average reflection rate of 25.8\% versus 37.5\% for hard questions.
    \item Within each difficulty class, correctness rate is not correlated with reflection rate, suggesting that model reflections may be redundant even on difficult questions.
    \item After intervention, reflection rates decrease across all difficulty classes, but harder questions still receive more reflections than easier ones.
\end{enumerate}

\subsection{Same-family model scale comparison}
\label{app:same_family}
To isolate the effect of model scale from architecture differences, we evaluate three models from the same family and distillation procedure: DeepSeek-R1-Distill-Qwen-1.5B, 7B, and 14B. All three are distilled from DeepSeek-R1 using identical training protocols, varying only in model capacity. We extract a separate reflection direction for each model from GSM8k traces and apply steering at $\lambda \in \{0, -0.48, -0.96\}$.

\paragraph{Reflection rates.}
\cref{tab:same_family_reflect} shows that reflection rates are similar across scales (19--23\%), indicating that the propensity to reflect is not strongly scale-dependent.

\begin{table}[!h]
\centering
\begin{tabular}{l|ccc}
\hline
Model & Total Steps & Reflect Steps & Reflect Rate \\
\hline
DS-Qwen-1.5B & 25{,}508 & 5{,}981 & 23.4\% \\
DS-Qwen-7B & 46{,}715 & 10{,}722 & 23.0\% \\
DS-Qwen-14B & 42{,}794 & 8{,}378 & 19.6\% \\
\hline
\end{tabular}
\caption{Reflection rates on GSM8k across the DeepSeek-R1-Distill-Qwen family (keyword heuristic).}
\label{tab:same_family_reflect}
\end{table}

\paragraph{Steering results.}
As shown in \cref{tab:same_family_steering}, larger models tolerate stronger reflection suppression with dramatically less accuracy cost. The efficiency ratio (token reduction per percentage point of accuracy drop) increases monotonically from 39$\times$ (1.5B) to 53$\times$ (7B) to 540$\times$ (14B), confirming that reflection redundancy scales with model capacity within a single architecture family.

\begin{table}[!h]
\centering
\begin{tabular}{l|c|cccc}
\hline
Model & $\lambda$ & Accuracy & Tokens & $\Delta$ Acc & $\Delta$ Tok \\
\hline
\multirow{3}{*}{DS-Qwen-1.5B} & 0.0 & 85.90\% & 1979 & --- & --- \\
& $-0.48$ & 84.89\% & 918 & $-1.01$pp & $-53.6\%$ \\
& $-0.96$ & 84.33\% & 781 & $-1.57$pp & $-60.5\%$ \\
\hline
\multirow{3}{*}{DS-Qwen-7B} & 0.0 & 92.90\% & 1517 & --- & --- \\
& $-0.48$ & 92.67\% & 992 & $-0.23$pp & $-34.6\%$ \\
& $-0.96$ & 92.01\% & 806 & $-0.89$pp & $-46.9\%$ \\
\hline
\multirow{3}{*}{DS-Qwen-14B} & 0.0 & 95.15\% & 1316 & --- & --- \\
& $-0.48$ & 95.15\% & 880 & $+0.00$pp & $-33.1\%$ \\
& $-0.96$ & 95.07\% & 748 & $-0.08$pp & $-43.2\%$ \\
\hline
\end{tabular}
\caption{\textbf{Same-family scale comparison on GSM8k.} Three DeepSeek-R1-Distill-Qwen models (1.5B/7B/14B), distilled with the same procedure, evaluated under identical steering settings. Larger models achieve similar token reduction with dramatically less accuracy cost, confirming that reflection redundancy scales with model capacity.}
\label{tab:same_family_steering}
\end{table}

\subsection{Token breakdown: reflection vs.\ non-reflection}
\label{app:token_breakdown}
To confirm that steering specifically targets reflection tokens rather than reducing reasoning output indiscriminately, we decompose the total token count into reflection and non-reflection components.
We tokenize all reasoning traces using the model's own tokenizer, split into steps at paragraph boundaries, classify each step as reflection or non-reflection via keyword matching, and count tokens per category.
Steps containing the final-answer phrase are excluded from both categories, consistent with the direction extraction procedure (\cref{sec:method}); the Reflect \% column is thus computed as reflection tokens divided by the sum of reflection and non-reflection tokens.

\cref{tab:token_breakdown} shows results for DS-Qwen-1.5B and DS-Qwen-7B on GSM8k.
At $\lambda{=}{-}0.96$, reflection tokens are reduced by 75--85\% while non-reflection tokens are reduced by only 34--44\%---a 2--3$\times$ disproportionate reduction.
The reflection-token fraction drops from 33--40\% of reasoning tokens to approximately 15\%, confirming that the intervention specifically suppresses reflection behavior rather than shortening reasoning uniformly.

\begin{table}[!h]
\centering
\begin{tabular}{l|c|ccc}
\hline
Model & $\lambda$ & Reflect Tok & Non-Reflect Tok & Reflect \% \\
\hline
\multirow{3}{*}{DS-Qwen-1.5B} & 0.0 & 775 & 1141 & 40.4\% \\
& $-0.48$ & 189\;{\scriptsize($-$75.6\%)} & 699\;{\scriptsize($-$38.7\%)} & 21.3\% \\
& $-0.96$ & 113\;{\scriptsize($-$85.4\%)} & 636\;{\scriptsize($-$44.3\%)} & 15.1\% \\
\hline
\multirow{3}{*}{DS-Qwen-7B} & 0.0 & 492 & 997 & 33.0\% \\
& $-0.48$ & 185\;{\scriptsize($-$62.3\%)} & 780\;{\scriptsize($-$21.8\%)} & 19.2\% \\
& $-0.96$ & 120\;{\scriptsize($-$75.5\%)} & 660\;{\scriptsize($-$33.8\%)} & 15.4\% \\
\hline
\end{tabular}
\caption{\textbf{Reflection vs.\ non-reflection token breakdown on GSM8k.} Reflection tokens are reduced 2--3$\times$ more than non-reflection tokens across both models and steering strengths, confirming that the intervention specifically targets reflection behavior.}
\label{tab:token_breakdown}
\end{table}

\subsection{Statistical reporting: standard deviations}
\label{app:std}
\cref{tab:std_gsm8k,tab:std_math500} report the standard deviations across 10 independent runs for all entries in the main results tables.
Accuracy standard deviations range from 0.20--1.18pp, confirming that the reported differences are well above run-to-run noise.

\begin{table}[!h]
\centering
\begin{tabular}{l|c|cc}
\hline
Model & $\lambda$ & Accuracy Std (pp) & Tokens Std \\
\hline
\multirow{4}{*}{DS-Llama-8B}
& $-0.96$ & 0.73 & 11 \\
& $-0.48$ & 0.46 & 15 \\
& $0$ & 0.46 & 45 \\
& $+0.48$ & 0.55 & 61 \\
\hline
\multirow{4}{*}{QwQ-32B}
& $-0.96$ & 0.22 & 13 \\
& $-0.48$ & 0.22 & 12 \\
& $0$ & 0.25 & 12 \\
& $+0.48$ & 0.27 & 32 \\
\hline
\multirow{4}{*}{DS-Qwen-14B}
& $-0.96$ & 0.20 & 7 \\
& $-0.48$ & 0.37 & 11 \\
& $0$ & 0.25 & 15 \\
& $+0.48$ & 0.32 & 74 \\
\hline
\end{tabular}
\caption{\textbf{Standard deviations for GSM8k results} (10 independent runs per setting).}
\label{tab:std_gsm8k}
\end{table}

\begin{table}[!h]
\centering
\begin{tabular}{l|c|cc}
\hline
Model & $\lambda$ & Accuracy Std (pp) & Tokens Std \\
\hline
\multirow{4}{*}{DS-Llama-8B}
& $-0.96$ & 0.92 & 115 \\
& $-0.48$ & 0.95 & 73 \\
& $0$ & 1.18 & 105 \\
& $+0.48$ & 0.51 & 132 \\
\hline
\multirow{4}{*}{QwQ-32B}
& $-0.96$ & 0.47 & 55 \\
& $-0.48$ & 0.64 & 42 \\
& $0$ & 0.37 & 41 \\
& $+0.48$ & 0.32 & 74 \\
\hline
\multirow{4}{*}{DS-Qwen-14B}
& $-0.96$ & 0.71 & 46 \\
& $-0.48$ & 0.60 & 66 \\
& $0$ & 0.44 & 81 \\
& $+0.48$ & 0.63 & 113 \\
\hline
\end{tabular}
\caption{\textbf{Standard deviations for MATH-500 results} (10 independent runs per setting).}
\label{tab:std_math500}
\end{table}

\subsection{Ablation study: impact of intervention layers}
\label{app:ablationLayer}
\begin{figure}[!h]
    \centering
    \includegraphics[width=0.6\linewidth]{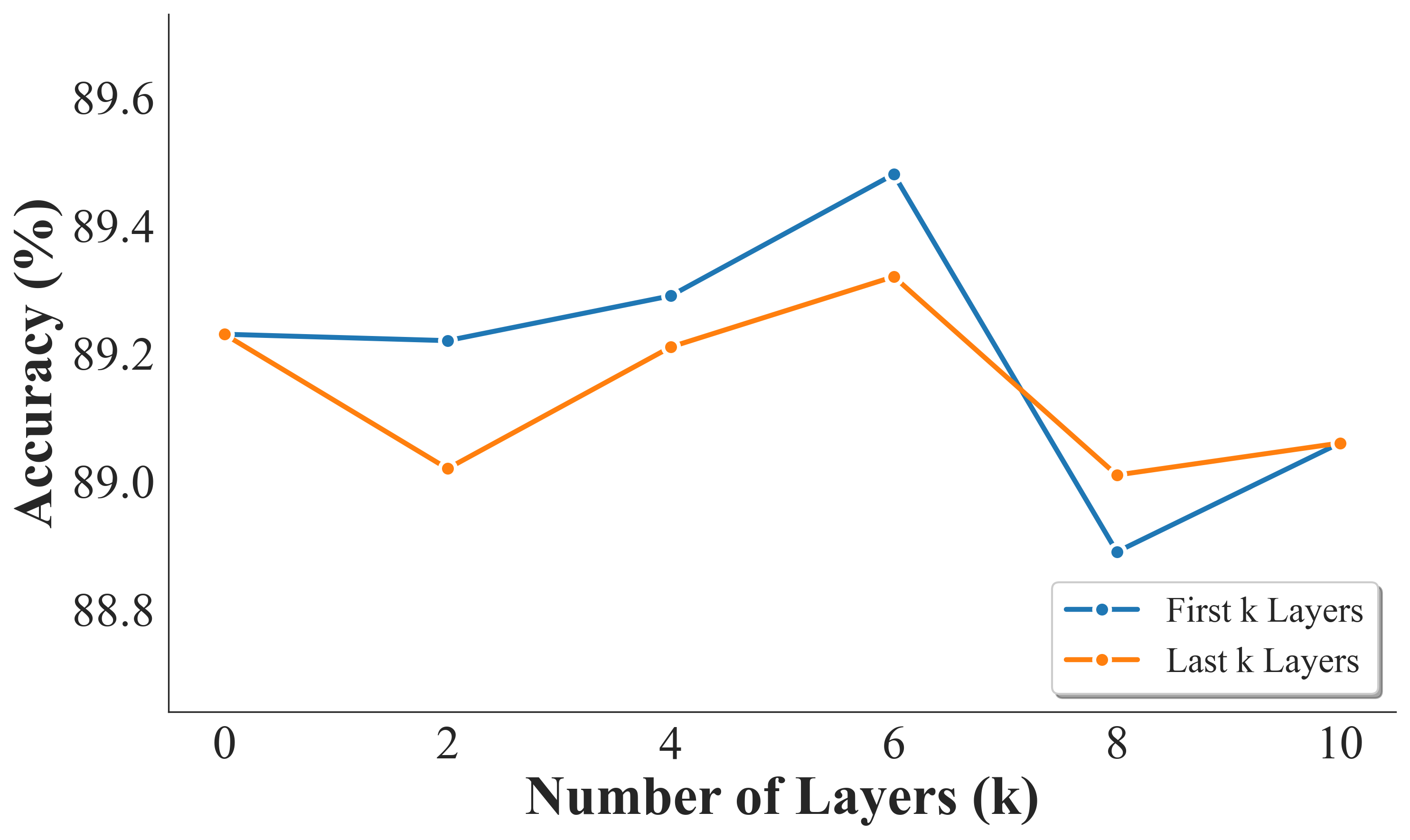}
    \caption{\textbf{Effect of applying interventions to different layers.} We vary the number of skipped layers at the bottom and top of the network at fixed strength $\lambda=-0.48$. Accuracy is highest when skipping the first and last six layers, which is adopted as the default.}
    \label{fig:layerAblation}
\end{figure}

We ablate the choice of which layers receive the intervention, testing two strategies: skipping the first $k$ layers and skipping the last $k$ layers. As shown in \cref{fig:layerAblation}, accuracy is highest when skipping both the first and last six layers, which we adopt as the default configuration.

\subsection{Ablation study: impact of components}
\label{app:ablationComponent}
\begin{figure}[!h]
    \centering
    \includegraphics[width=0.98\linewidth]{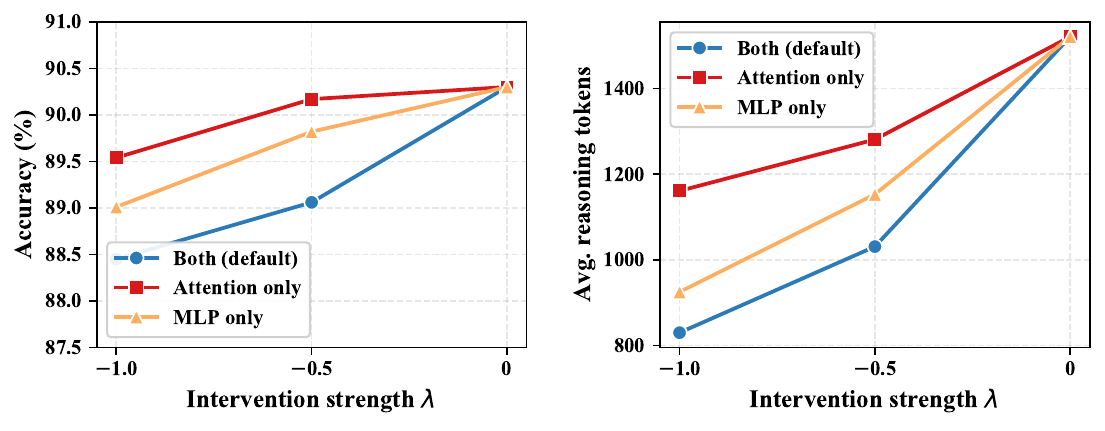}
    \caption{Component ablation: accuracy and reasoning token usage when steering is applied to self-attention only, MLP only, or both (default). Results on GSM8k with DeepSeek-R1-Distill-Llama-8B.}
    \label{fig:component_ablation}
\end{figure}

We ablate which model component receives the steering intervention: self-attention only, MLP only, or both (default). All experiments use DeepSeek-R1-Distill-Llama-8B on GSM8k with 3 samples per question. \cref{fig:component_ablation} shows accuracy and token usage at strengths $\lambda \in \{0, -0.5, -1.0\}$. We find that
\textbf{combining both components yields the strongest token reduction.} At $\lambda=-1.0$, the default setting reduces tokens to 830, compared to 1{,}161 for attention-only and 925 for MLP-only, while maintaining comparable accuracy.

\clearpage
\section{Validation of the reflection keyword heuristic}
\label{app:reflection-audit}
The reflection direction extraction (\cref{subsec:identify}) relies on a keyword-based heuristic that labels a reasoning step as \emph{reflective} if it contains any of the keywords: \texttt{wait}, \texttt{let me check}, \texttt{double-check}, or \texttt{alternatively} (case-insensitive).
To assess the quality of this heuristic, we compare its labels against an independent LLM judge on unsteered ($\lambda=0$) GSM8k reasoning traces.

\paragraph{Setup.}
We use Qwen3-32B~\citep{yang2025qwen3} as the judge, served via vLLM with 4-way tensor parallelism and greedy decoding ($\text{temperature}=0$).
Each reasoning trace is segmented into steps by splitting on the ``\textbackslash n\textbackslash n'' delimiter.
Steps are grouped into non-overlapping chunks of $k{=}4$ target steps, each preceded by $c{=}2$ context steps from the prior chunk for local coherence.
We audit 1{,}000 DeepSeek-R1-Distill-Llama-8B reasoning traces.

The judge is instructed to label whether each target step is the \emph{start} of a new reflection span---directly mirroring the onset signal used by our stepwise steering intervention.
The full system prompt is shown below:

\begin{quote}
\small\ttfamily
You are an expert annotator for mathematical reasoning traces.
Your task is to identify whether each TARGET step is the START of a self-reflection span.

Definition of self-reflection start step:
Label 1 only if the step is the first step where reflection behavior begins
(e.g., doubt/check/correction/alternative-path trigger starts at this step).

Label 0 for:
continuation steps inside an already-started reflection span,
normal forward derivation,
routine arithmetic/logical progress,
final answer formatting without re-checking intent.

Important rules:
Only label TARGET steps, not CONTEXT steps.
Use CONTEXT to decide whether reflection already started before a TARGET step.
If reflection is already ongoing in CONTEXT, early TARGET continuation steps must be 0.
\end{quote}

\noindent The judge returns a structured JSON response with a binary reflection label, confidence score, and brief rationale for each target step.

\paragraph{Results.}
\cref{tab:reflection-audit} summarizes the audit over 28{,}413 labeled steps across 9{,}306 chunks.
The overall agreement between the heuristic and the judge is 87.3\%, driven by the large number of non-reflective steps on which both methods concur.
The per-class metrics show that the keyword heuristic achieves 64.8\% precision but only 59.0\% recall, indicating that the heuristic is a conservative detector---most steps it flags are genuine reflection onsets, but it misses a substantial fraction that the judge identifies.

\begin{table}[t]
\centering
\caption{Reflection heuristic validation against an LLM judge (Qwen3-32B) on DeepSeek-R1-Distill-Llama-8B. Precision and recall treat the judge labels as reference.}
\label{tab:reflection-audit}
\small
\begin{tabular}{@{}lr@{}}
\toprule
Labeled steps        & 28{,}413 \\
Chunks evaluated     & 9{,}306 \\
\midrule
Heuristic pos.\ rate & 15.8\% \\
Judge pos.\ rate     & 17.3\% \\
\midrule
TP / FP / FN / TN    & 2{,}899 / 1{,}577 / 2{,}018 / 21{,}919 \\
\midrule
Agreement             & 87.3\% \\
Precision             & 64.8\% \\
Recall                & 59.0\% \\
$F_1$                 & 61.7\% \\
\bottomrule
\end{tabular}
\end{table}

Inspection of disagreement cases reveals two systematic patterns:
\begin{enumerate}
    \item \textbf{False negatives} (heuristic misses) arise when the model initiates reflection with phrases not covered by the keyword list, such as \textit{``Hold on, is that right?''} or implicit re-checking without a lexical trigger.
    \item \textbf{False positives} (heuristic over-detections) occur when a keyword appears inside an already-active reflection span; the heuristic flags every keyword occurrence, whereas the judge correctly restricts the positive label to the span's first step.
\end{enumerate}
Despite the heuristic's imperfect recall, the extracted steering directions remain effective (\cref{tab:main_results}).
We attribute this to the contrastive nature of the mean-difference extraction: even a noisy partition of steps into reflective vs.\ non-reflective sets yields a meaningful direction, because the true positives dominate the reflective centroid while the missed steps only slightly dilute the non-reflective centroid.

\paragraph{Direction robustness: keyword vs.\ LLM-judge labels.}
To directly quantify the impact of label noise on the extracted direction, we re-extract the reflection direction using the LLM judge's labels instead of the keyword heuristic.
Using all 1{,}319 GSM8k traces from DeepSeek-R1-Distill-Llama-8B (52{,}346 total steps, of which 8{,}427 are labeled by the judge), we compute the mean-difference direction from the judge-labeled reflective vs.\ non-reflective partitions and compare it with the keyword-based direction used in all main experiments.

\cref{tab:direction_comparison} reports the per-component cosine similarity between the two directions across all 64 sub-components (32 layers $\times$ \{attention, MLP\}).
The mean cosine similarity is 0.81, indicating that both labeling methods recover largely the same linear direction.

\begin{table}[!h]
\centering
\begin{tabular}{l|cccc}
\hline
& Mean & Std & Min & Max \\
\hline
Cosine similarity & \textbf{0.81} & 0.045 & 0.71 & 0.94 \\
\hline
\end{tabular}
\caption{Cosine similarity between keyword-based and LLM-judge-based reflection directions across all 64 sub-components (DS-Llama-8B, GSM8k).}
\label{tab:direction_comparison}
\end{table}

To confirm this translates to equivalent downstream performance, we run steering with the LLM-judge direction at $\lambda{=}{-}0.48$ on both GSM8k and MATH-500 (\cref{tab:judge_steering}).
The two directions produce nearly identical results: accuracy differences are within 1.6pp and token counts are comparable, confirming that the mean-difference extraction is inherently robust to label noise.

\begin{table}[!h]
\centering
\begin{tabular}{l|l|cc}
\hline
Dataset & Direction source & Accuracy & Tokens \\
\hline
\multirow{2}{*}{GSM8k} & Keyword (paper) & 89.46\% & 1033 \\
& LLM-judge & 88.63\% & 1041 \\
\hline
\multirow{2}{*}{MATH-500} & Keyword (paper) & 84.46\% & 3124 \\
& LLM-judge & 82.87\% & 2765 \\
\hline
\end{tabular}
\caption{Steering comparison using keyword-based vs.\ LLM-judge-based reflection directions (DS-Llama-8B, $\lambda{=}{-}0.48$). Both directions produce equivalent steering outcomes, confirming robustness to label noise in direction extraction.}
\label{tab:judge_steering}
\end{table}

\clearpage
\section{Computational resources}
\label{app:compute}
The main experiments were conducted on 8 NVIDIA RTX 6000 Ada GPUs for approximately 200 hours.

\section{LLM Disclosure}
\label{app:llm-disclosure}
This work uses Qwen3-32B~\citep{yang2025qwen3} as an LLM-based judge to validate the reflection keyword heuristic (\cref{app:reflection-audit}), as well as the DeepSeek-R1-Distill family and QwQ-32B in the main experiments. The writing of this paper was polished with the assistance of an LLM.

\end{document}